\documentclass[11pt,a4paper]{article}
\usepackage[T1]{fontenc}
\usepackage[nomath]{lmodern}
\usepackage[htt]{hyphenat}

\providecommand{\ReportTitle}{Qiushi Engine on AstaBench\\E2E-Bench-Hard}

\providecommand{\ReportAuthor}{%
  Wenhao Li,\quad Shuxing Yang,\quad Fujia Chen,\quad Jincheng Mi,\\[0.15em]
  Yuang Pan,\quad Rui Zhao,\quad Zichen Li,\quad Junyao Wu,\\[0.15em]
  Shenzhan Hong,\quad Yaqi Li,\quad Yize Wang,\quad Kaihao Zhu,\\[0.15em]
  Taowen Deng,\quad Junjie Yang,\quad Hongsheng Chen,\quad Yihao Yang\textsuperscript{*}}
\providecommand{\ReportOrganization}{Qiushi Engine Team, College of Information Science and Electronic Engineering\\Zhejiang University}
\providecommand{\ReportCorrespondence}{\textsuperscript{*}Corresponding author: \href{mailto:yangyihao@zju.edu.cn}{\nolinkurl{yangyihao@zju.edu.cn}}}

\providecommand{\ReportVersion}{v2.0}
\providecommand{\ReportDate}{September 2026}

\usepackage{amsmath,amssymb}
\usepackage{siunitx}
\usepackage{graphicx}
\usepackage[dvipsnames,table]{xcolor}
\usepackage{tikz}
\usepackage{pgfplots}\pgfplotsset{compat=1.18}
\graphicspath{{figures/}}

\usepackage{booktabs,longtable,array,tabularx,multirow,makecell,threeparttable}
\usepackage{pdflscape}

\newif\ifQiushiHasCitations
\let\QiushiOriginalCite\cite
\renewcommand{\cite}{\global\QiushiHasCitationstrue\QiushiOriginalCite}
\let\QiushiOriginalNocite\nocite
\renewcommand{\nocite}{\global\QiushiHasCitationstrue\QiushiOriginalNocite}

\usepackage[inner=2.25cm,outer=2.25cm,top=2.25cm,bottom=2.35cm,headheight=24pt,footskip=28pt]{geometry}
\usepackage{setspace}
\usepackage{microtype}
\usepackage{caption,subcaption}
\usepackage{enumitem}\setlist{itemsep=0.3em,topsep=0.5em,parsep=0.1em}
\usepackage{float}
\usepackage{afterpage}
\usepackage{listings}

\definecolor{ReportPrimary}{HTML}{1F3A5F}
\definecolor{ReportAccent}{HTML}{8A5A2B}
\definecolor{ReportMuted}{HTML}{6F7782}
\definecolor{ReportLight}{HTML}{F4F7FA}
\definecolor{ReportRule}{HTML}{D6DDE5}
\definecolor{ReportFinding}{HTML}{2D5A87}
\definecolor{ReportRecommend}{HTML}{2D7A4F}
\definecolor{ReportRisk}{HTML}{B7472A}
\definecolor{ReportInsight}{HTML}{6B4E94}

\usepackage{titlesec}
\titleformat{\section}
  {\Large\bfseries\color{black}}
  {\thesection}{0.6em}{}
\titlespacing*{\section}{0pt}{1.55em}{0.65em}
\titleformat{\subsection}{\large\bfseries\color{black}}{\thesubsection}{0.5em}{}
\titlespacing*{\subsection}{0pt}{1.2em}{0.45em}
\titleformat{\subsubsection}{\normalsize\bfseries}{\thesubsubsection}{0.5em}{}

\usepackage{fancyhdr}
\renewcommand{\headrulewidth}{0.4pt}
\renewcommand{\footrulewidth}{0.4pt}
\renewcommand{\headrule}{\hbox to\headwidth{\color{black!55}\leaders\hrule height \headrulewidth\hfill}}
\renewcommand{\footrule}{\hbox to\headwidth{\color{ReportRule}\leaders\hrule height \footrulewidth\hfill}}
\fancypagestyle{plain}{%
  \fancyhf{}%
  \fancyhead[L]{\footnotesize\color{black}Qiushi Engine on AstaBench E2E-Bench-Hard}%
  \fancyfoot[R]{\small\thepage}%
  \renewcommand{\headrulewidth}{0.4pt}%
  \renewcommand{\footrulewidth}{0.4pt}%
  \renewcommand{\headrule}{\hbox to\headwidth{\color{black!55}\leaders\hrule height \headrulewidth\hfill}}%
  \renewcommand{\footrule}{\hbox to\headwidth{\color{ReportRule}\leaders\hrule height \footrulewidth\hfill}}%
}
\fancypagestyle{firstpage}{\fancyhf{}\fancyfoot[L]{\footnotesize\color{black!70}Wenhao Li et al.}\fancyfoot[R]{\small\thepage}\renewcommand{\headrulewidth}{0pt}\renewcommand{\footrulewidth}{0.4pt}\renewcommand{\footrule}{\hbox to\headwidth{\color{black!55}\leaders\hrule height \footrulewidth\hfill}}}

\rowcolors{2}{ReportLight}{white}

\usepackage{xurl}
\usepackage{hyperref}
\hypersetup{colorlinks=true,linkcolor=ReportPrimary,citecolor=ReportAccent,urlcolor=ReportPrimary,
            pdftitle={Qiushi Engine on AstaBench E2E-Bench-Hard},
            pdfauthor={Wenhao Li; Shuxing Yang; Fujia Chen; Jincheng Mi; Yuang Pan; Rui Zhao; Zichen Li; Junyao Wu; Shenzhan Hong; Yaqi Li; Yize Wang; Kaihao Zhu; Taowen Deng; Junjie Yang; Hongsheng Chen; Yihao Yang}}
\usepackage{bookmark}
\usepackage{cleveref}

\DeclareUrlCommand\qpath{\urlstyle{tt}\small}
\DeclareUrlCommand\qurl{\urlstyle{same}\small}

\newcolumntype{Y}{>{\raggedright\arraybackslash\hspace{0pt}}X}
\newcolumntype{L}[1]{>{\raggedright\arraybackslash\hspace{0pt}}p{#1}}
\usepackage[most]{tcolorbox}
\tcbuselibrary{breakable,skins}
\newtcolorbox{keyfindingbox}[1][Key Finding]{
  breakable, enhanced,
  colback=white, colframe=black!55,
  coltitle=black, fonttitle=\bfseries,
  title=#1, boxrule=0.45pt, arc=0mm,
  left=10pt,right=10pt,top=8pt,bottom=8pt}
\newtcolorbox{insightbox}[1][Insight]{
  breakable, enhanced,
  colback=white, colframe=black!55,
  coltitle=black, fonttitle=\bfseries,
  title=#1, boxrule=0.45pt, arc=0mm,
  left=10pt,right=10pt,top=8pt,bottom=8pt}

\newcommand{\QiushiBrand}{%
  \IfFileExists{assets/qiushi_engine_logo_original.png}{%
    \includegraphics[width=5.20cm,keepaspectratio]{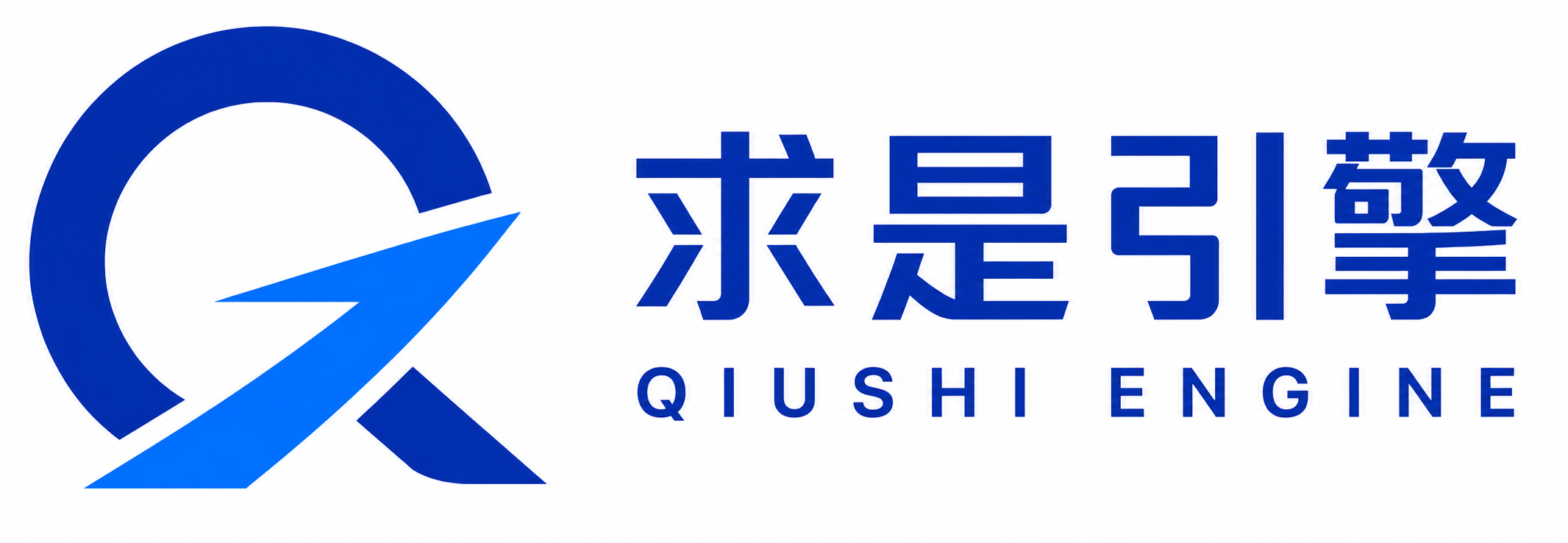}%
  }{%
    \PackageError{qiushi-report}{Missing original logo file assets/qiushi_engine_logo_original.png}{Place the user-supplied PNG in source/assets/ and compile again.}%
  }%
}

\newcommand{\MakeQiushiOpening}{%
  \noindent\QiushiBrand\par
  \vspace{0.35em}
  \noindent{\color{black!55}\rule{\textwidth}{0.5pt}}\par
  \vspace{1.0em}
  {\fontsize{24}{31}\selectfont\bfseries\color{black}\ReportTitle\par}
  \vspace{0.8em}
  {\noindent\large\bfseries\ReportAuthor\par}
  \vspace{0.3em}
  {\noindent\small\ReportOrganization\par}
  \vspace{0.3em}
  {\noindent\small\ReportCorrespondence\par}
  \vspace{0.3em}
  {\small\color{black!65}\ReportDate\quad \ReportVersion\par}
  \vspace{1.25em}
}

\begin{document}

\setcounter{page}{1}

\thispagestyle{firstpage}
\MakeQiushiOpening


\noindent\begin{minipage}{\textwidth}
\small\bfseries
This report analyzes Qiushi Engine v0.8 across all 40 test tasks in AstaBench E2E-Bench-Hard, a benchmark that requires autonomous agents to carry a research question through experimental design, code implementation, actual execution, result analysis, and report delivery. Qiushi Engine is model-configurable; this evaluation selected DeepSeek \texttt{deepseek-v4pro-preview} as the model backend. The official AstaBench leaderboard records a score of 0.816 and an average benchmark cost of \$15.209 per task, while the full-precision local recomputation is $81.59 \pm 1.87$. Four tasks satisfied every rubric item, yielding a full-task completion rate of $4/40=10\%$---7 percentage points above, and about 3.3 times, the approximately 3\% best rate reported for AstaBench's official agents. Across 507 required rubric items, 416 were satisfied (82.1\%). Official scoring archives and 40 Meta-Trace records show sustained production and verification of reports, code, and experimental artifacts; the principal gaps lie in repeated runs, external dependencies, specified metrics, and ablation studies. The report explains the benchmark, system workflow, aggregate results, representative cases, and limits of interpretation.
\end{minipage}

\vspace{1.0em}

\section{Introduction}
\label{sec:intro}

\subsection{The End-to-End Scientific Discovery Challenge}

The question of whether LLM-based agents can perform genuine scientific research---not merely answer questions, write code, or retrieve papers, but carry out full research cycles that produce real experimental evidence---has become a central challenge in AI evaluation. Unlike narrow coding benchmarks where a single correct output suffices, end-to-end scientific discovery requires an agent to interpret a research question, design an appropriate experimental methodology, implement working code, execute experiments on real data, analyze results with statistical rigor, and produce a coherent technical report whose claims are supported by the generated code and artifacts.

This challenge is qualitatively different from the tasks that existing LLM benchmarks measure. HumanEval, MBPP, and DS-1000 test whether a model can produce executable code from a specification~\cite{chen2021humaneval,austin2021mbpp,lai2022ds1000}. MMLU and GPQA test broad academic knowledge and difficult scientific question answering~\cite{hendrycks2020mmlu,rein2023gpqa}. SWE-bench and SUPER move toward realistic repository-level issue resolution, environment setup, and execution of research code~\cite{jimenez2023swebench,bogin2024super}. End-to-end scientific discovery sits at the intersection of these capabilities while adding experiment design, evidence generation, result interpretation, and coherent reporting.

The difficulty of this challenge has been confirmed empirically. Studies of frontier AI agents on research-like tasks have found that while agents can often complete individual engineering steps, the overall quality of research outputs remains far below expert standards. The compounding nature of multi-step research means that even moderate per-step failure rates produce very low end-to-end completion rates.

\subsection{AstaBench and E2E-Bench-Hard}

AstaBench~\cite{bragg2025astabench} is a broad benchmark suite designed to measure scientific-agent capabilities across 11~benchmarks and four categories---Literature Understanding, Code \& Execution, Data Analysis, and End-to-End Discovery---with over 2{,}400 problems in total. The benchmarks are designed to remain reproducible as science progresses, using date-limited access to the scientific literature and standardized computational environments.

Among the 11~benchmarks, E2E-Bench and E2E-Bench-Hard target End-to-End Discovery. Each task requires the agent to complete a full AI/NLP research pipeline, and the final score reflects the average proportion of task-specific rubric items jointly supported by the generated report, code, and artifacts.

Research-agent evaluation now forms a progression of increasingly complete tasks. AgentBench measures long-horizon decision making in interactive environments~\cite{liu2023agentbench}. MLAgentBench and MLE-bench focus on machine-learning experimentation, data preparation, model training, and iterative improvement~\cite{huang2023mlagentbench,chan2024mlebench}. CORE-Bench and PaperBench evaluate computational reproducibility and paper-level research replication~\cite{siegel2024corebench,starace2025paperbench}. ScienceAgentBench tests data-driven scientific tasks grounded in peer-reviewed papers~\cite{chen2024scienceagentbench}, while the AI Scientist demonstrates an automated pipeline spanning ideation, coding, experimentation, paper writing, and simulated review~\cite{lu2024aiscientist}. AstaBench is distinctive within this progression because its unsimplified research prompts and evidence-linked rubrics directly evaluate the complete path from task definition to verifiable research deliverables.

E2E-Bench-Hard is distinguished from E2E-Bench by its task generation method. While E2E-Bench tasks undergo a simplification step to ensure tractability, E2E-Bench-Hard tasks are generated by HypER (Hypothesis-driven Exploration of Research trends), which identifies research trends from highly cited ACL papers, proposes underexplored research directions, refines them using relevant paper excerpts, and sends them through automatic ranking and expert human review---without simplification. The result is a set of 50~tasks (40~test, 10~validation) that are genuinely hard: they typically require 10--15 distinct research steps, and AstaBench reports that even when individual step completion rates are reasonable (up to ${\sim}70\%$), the probability of completing all required steps in a single task remains near zero for all published agents due to compounding failures.

\subsection{Scope and Structure of This Report}

This report analyzes the performance of Qiushi Engine v0.8 on all 40 E2E-Bench-Hard test tasks, providing a complete evidence chain from the benchmark's scoring mechanism to the observed per-sample outcomes. The analysis follows a continuous argument:

\begin{itemize}
\item \Cref{sec:benchmark} describes precisely how AstaBench designs and scores E2E-Bench-Hard tasks, including the task structure, rubric mechanism, and what the resulting score measures.
\item \Cref{sec:architecture} introduces the Qiushi Engine architecture relevant to understanding the observed process signatures.
\item \Cref{sec:submission} documents the submission protocol, model identity evidence, official cost, and resource usage.
\item \Cref{sec:aggregate} presents the aggregate results and their distributional structure.
\item \Cref{sec:process} is the central analytical section: it connects sample-level rubric outcomes to Qiushi's recorded workflow, explaining which observed mechanisms accompanied high scores and which experimental or artifact gaps caused score losses, with detailed case studies across the full score range.
\item \Cref{sec:comparison} places the result in the context of the official AstaBench leaderboard, using the displayed official score and cost fields.
\item \Cref{sec:discussion} discusses the implications, limitations, and improvement opportunities.
\item The appendices provide the full 40-sample evidence table, rubric inventory, archive contents, and data processing methodology.
\end{itemize}

\section{How AstaBench E2E-Bench-Hard Scores Research}
\label{sec:benchmark}

Understanding the Qiushi result requires understanding precisely what the score measures. AstaBench's End-to-End Discovery scoring is not a binary pass/fail, nor a single-number quality rating. It is a structured rubric-item evaluation that examines whether each required facet of a valid research result is supported by the agent's actual outputs.

\subsection{Task Structure and Specification}

Each E2E-Bench-Hard task provides the agent with a detailed research specification containing the following fields:

\begin{itemize}
\item \textbf{Task name and short description}: A concise statement of the research direction, e.g., ``Dynamic Commonsense Integration'' or ``Anticipatory Ensemble for Medication Extraction.''
\item \textbf{Hypothesis to explore}: The scientific hypothesis that the experiment should test, framed as a testable claim about the relationship between an independent variable (the proposed method or integration) and a dependent variable (a measurable outcome).
\item \textbf{Independent and dependent variables}: Explicit identification of what is being varied and what is being measured.
\item \textbf{Comparison groups}: The experimental conditions that should be compared, typically including the proposed method, relevant baselines, and ablation conditions.
\item \textbf{Baseline or control conditions}: Specific control systems or configurations that provide reference points for the experimental comparison.
\item \textbf{Measurement method}: The metrics, evaluation procedures, and statistical tests required to assess the hypothesis.
\end{itemize}

The specification is detailed enough to define the experimental scope but not so prescriptive as to reduce the task to code completion---the agent must still make numerous design decisions about implementation architecture, data handling, model selection and configuration, evaluation pipeline design, and reporting structure.

For example, the task ``Dynamic Commonsense Integration'' (idea\_5) asks the agent to integrate dynamic knowledge selection from ConceptNet and ATOMIC using a context-aware emotional graph attention mechanism, compare it against static integration baselines on a dialogue dataset, and measure F1-scores for emotion classification with statistical significance testing and multiple experimental runs. The task ``Anticipatory Ensemble for Medication Extraction'' (idea\_16) asks for combining anticipatory prompts with ensemble approaches for clinical medication attribute extraction, comparing against heuristic and chain-of-thought prompts, with specific requirements for GPT-3.5 and LLaMA-2 model integration, precision/recall metrics, and model-size comparison.

\subsection{Task Generation via HypER}

The tasks in E2E-Bench-Hard are generated through HypER, a pipeline that:
\begin{enumerate}
\item Identifies research trends by analyzing clusters of highly cited ACL papers.
\item Proposes underexplored research directions within those trends using an LLM.
\item Refines proposed ideas using relevant excerpts from the identified papers.
\item Ranks the refined ideas automatically and submits the top candidates to expert human raters.
\item Retains only those tasks that pass expert review for scientific validity and experimental feasibility.
\end{enumerate}

Critically, E2E-Bench-Hard omits the simplification step that E2E-Bench applies. In E2E-Bench, after HypER generates a task, a separate simplification pass reduces the number of required components and experimental conditions to ensure the task is tractable within moderate compute budgets. E2E-Bench-Hard skips this step, preserving the full complexity of the original research specification. This makes the tasks harder not by requiring more sophisticated algorithms, but by requiring more complete experimental coverage---more comparison conditions, more metrics, more statistical analyses, and more ablation studies.

\subsection{Rubric Design}

Each task comes with a task-specific rubric: a list of criteria, each phrased as a yes/no question about whether a specific research facet was accomplished. These rubrics were manually checked and revised by human annotators to ensure that each item is: (1)~necessary for a valid investigation of the stated hypothesis, (2)~clearly defined enough to be evaluable from the agent's outputs, and (3)~distinguishable from other items in the rubric.

The rubric items span several categories of research activity:

\begin{itemize}
\item \textbf{Implementation items}: Did the agent implement the proposed method, baselines, integration mechanisms, and supporting modules?
\item \textbf{Data items}: Did the agent obtain or construct the required datasets and prepare them appropriately?
\item \textbf{Measurement items}: Did the agent compute the specified metrics (accuracy, F1, BLEU, BERTScore, etc.)?
\item \textbf{Statistical items}: Did the agent perform statistical significance testing, multiple independent runs, or meta-analysis?
\item \textbf{Analysis items}: Did the agent conduct ablation studies, sensitivity analyses, error analyses, or qualitative assessments?
\item \textbf{Model-specific items}: Did the agent use the specified model architectures or backbone models?
\item \textbf{Reporting items}: Did the agent produce documentation, prompt designs, or method descriptions?
\end{itemize}

The number of rubric items varies across tasks. In the 40~test samples analyzed here, the count of scored rubric items ranges from 8 to 17 (Q1~=~11, median~=~12.5, Q3~=~14), with a mean of 12.7 items per sample. Each task also includes optional rubric items (typically 5--8 additional items) that are listed in the task target but excluded from scoring.

\subsection{Scoring Mechanism}

The scoring proceeds in three steps:

\textbf{Step 1: Artifact extraction.} The AstaBench E2E scorer extracts the agent's three output artifact classes: the generated \textit{report} (a technical document describing the experiment and results), the generated \textit{code} (implementation files), and the generated \textit{artifacts} (datasets, result files, figures, model outputs, or other research products).

\textbf{Step 2: Per-item evaluation.} For each rubric item, an LLM-as-judge (in this evaluation, \texttt{claude-sonnet-4-6}) evaluates whether the criterion is met by examining all three artifact classes together. The judge uses cross-artifact consistency: a claim in the report is not sufficient if the code does not implement it or the artifacts do not contain the expected evidence. Conversely, code that implements a feature but is not reflected in the report or artifacts may also fail. The judge produces a binary verdict (1 if met, 0 if not) for each item.

\textbf{Step 3: Score aggregation.} The sample score is the arithmetic mean of the binary item scores. A sample with 12 scored items, of which 10 are met, receives a score of $10/12 = 0.833$. The task-level score is the arithmetic mean of all 40 sample scores, with standard error computed across samples.

This design has an important consequence for interpretation: a score of 0.80 does not mean ``80\% quality.'' It means that 80\% of the required research facets---each representing a concrete implementation, measurement, comparison, or analysis step---were judged as supported by the combined report, code, and artifact evidence. The remaining 20\% represent specific, identifiable gaps in the research output.

\subsection{Why E2E-Bench-Hard Is Difficult}

The compounding nature of end-to-end research makes these tasks qualitatively different from isolated coding or QA problems. A typical task requires the agent to: (1)~download or construct appropriate datasets, (2)~implement the proposed method and baselines, (3)~run experiments with proper controls, (4)~compute specified metrics, (5)~perform statistical testing, (6)~conduct ablation or sensitivity analyses, (7)~produce error analysis or qualitative assessment, and (8)~write a coherent report that accurately reflects the code and data evidence. Failure at any step can cascade: if a required model cannot be loaded, multiple downstream rubric items---its evaluation metrics, statistical comparisons, and ablation studies---may all score zero.

AstaBench's own analysis confirms this compounding difficulty. Table~20 of the AstaBench paper reports that the overall end-to-end task completion rate (all required steps completed successfully) is near zero for all published agents: Asta Panda with Claude Sonnet~4 completes 3\% of E2E-Bench-Hard tasks fully, Asta CodeScientist completes 3\%, and all other agents complete 0\%, even when the average per-step completion rate reaches ${\sim}70\%$. This stark contrast between step-level and task-level success rates underscores the compounding challenge that E2E-Bench-Hard is designed to measure.

\section{Qiushi Engine Architecture}
\label{sec:architecture}

Qiushi Engine is an autonomous research system with a configurable LLM backend. For this evaluation, the selected backend was DeepSeek \texttt{deepseek-v4pro-preview}. This section describes only the public workflow elements needed to interpret the E2E-Bench-Hard evidence; it is not a complete system specification.

\subsection{Research Functions}

The public workflow can be described through four functions without exposing internal routing details:

\begin{itemize}
\item \textbf{Plan}: interpret the task, identify required datasets, models, comparisons, and measurements, and choose an experimental route.
\item \textbf{Build}: implement data processing, methods, baselines, metrics, and report-ready artifacts.
\item \textbf{Run and repair}: execute experiments, inspect outputs, diagnose failures, and revise code or configuration when needed.
\item \textbf{Verify and communicate}: check consistency among claims, code, and generated artifacts, then assemble the final evidence package.
\end{itemize}

\subsection{Meta-Trace Memory System}

The workflow records each substantive step in the Meta-Trace memory system. Each recorded step contains two components:

\begin{itemize}
\item \textbf{Reasoning record}: a summary of the scientific judgment at that step---what was learned, what changed, what evidence was inspected, and what remains unresolved.
\item \textbf{Structured handoff}: the next task, current state (Process, Repair, Verify, Plan, Review, Reflect, Evaluate, or End), and research phase.
\end{itemize}

This creates a persistent, inspectable record of the research process. The Meta-Trace serves two functions: it provides long-horizon memory so that later agents can recover decisions and evidence from earlier steps, and it provides the process evidence analyzed in this report.

\subsection{Research Phases}

Work proceeds through three non-linear phases:

\textbf{Explore} builds understanding of the task, reads relevant sources, forms a research plan, and evaluates alternative approaches. In E2E-Bench-Hard tasks, Explore typically involves parsing the task specification, identifying which datasets and models are required, and deciding the implementation architecture.

\textbf{Execute} implements, runs, and verifies experiments. This is where the bulk of the work happens: writing code, running scripts, capturing outputs, diagnosing errors, repairing failures, and iterating until the experimental results are satisfactory.

\textbf{Express} synthesizes results into a final report and artifact bundle. This phase packages the research outputs into the JSON structure that the AstaBench scorer expects (report, code, trace, and artifacts).

Phase transitions are driven by the evidence state, not by a fixed schedule---the workflow can return to Explore after discovering that an initial approach fails, or re-enter Execute to repair a broken experiment during the Express phase. This flexibility is visible in the Meta-Trace state distributions: the 184~Repair states across 40 runs indicate that the system frequently detects and fixes problems during execution.

\subsection{Digital Workspace}

The digital workspace maintains research files---scripts, data, figures, notes, and deliverables---as inspectable assets. Search, history recovery, execution, and verification capabilities are invoked as the task requires. This artifact-centered design matters because the AstaBench scorer evaluates evidence in the report, code, and generated files rather than the plausibility of prose alone.

\subsection{Why This Architecture Matters for E2E-Bench-Hard}

The workflow is directly relevant to the scoring outcome. AstaBench evaluates three artifact classes (report, code, and generated artifacts) and checks whether they support one another. Qiushi's build, execution, repair, and verification functions produce these evidence classes as part of the same research loop. The final checks ask whether reported claims are supported by actual code and output files, which closely matches the benchmark's scoring target.

This does not guarantee success. The process can fail to satisfy rubric items when: a required external model is unavailable or cannot be loaded in the execution environment; the task demands more experimental runs than the step budget permits; the implementation misinterprets a technical requirement; or ablation and sensitivity analyses are deprioritized in favor of core implementation under time pressure.

\section{Submission Protocol and Model Identity}
\label{sec:submission}

Precise documentation of the model identity and submission path is essential for reproducibility and comparison. This section separates the three distinct model roles that appear in the submission archive.

\subsection{Three-Role Model Identity}

\begin{table}[htbp]
\centering
\caption{Model identity separation in the Qiushi submission archive. Three model identifiers appear in the archive serving fundamentally different roles; conflating them would misrepresent the evaluation.}
\label{tab:model-identity}
\begin{tabularx}{\textwidth}{L{3.2cm}L{3.8cm}Y}
\toprule
\textbf{Role} & \textbf{Identifier} & \textbf{Scope and Evidence} \\
\midrule
Qiushi agent model & \texttt{deepseek-v4pro-preview} & All 40 test runs. UI label: \texttt{deepseek-v4-pro (952K ctx, 48K out, premium; text)}. Confirmed by submitter in \texttt{test\_model\_configuration.json}. Qiushi's model backend is configurable; DeepSeek was selected for this evaluation. \\
Inspect replay label & \texttt{mockllm/model} & Offline replay wrapper used to package cached outputs into Inspect format. This label appears in replay-mode run objects, not in the archive header metadata. It does not represent the model that generated the original research outputs. \\
Rubric judge model & \texttt{claude-sonnet-4-6} & Used by the AstaBench E2E scorer to evaluate rubric items. Part of the scoring infrastructure, not the Qiushi agent. This local evaluation used \texttt{claude-sonnet-4-6} as the rubric judge. \\
\bottomrule
\end{tabularx}
\end{table}

\Cref{tab:model-identity} makes the separation explicit. The distinction matters because the Inspect archive header records the model as \texttt{deepseek/deepseek-v4pro-preview} (the Inspect/task metadata label) while the actual LLM that generated all research outputs was DeepSeek \texttt{deepseek-v4pro-preview}, as confirmed in the submission's \texttt{test\_model\_configuration.json}. The configuration record also states: ``Qiushi Engine is not DeepSeek-based as a product constraint; its model backend is selectable. DeepSeek was the backend selected for this evaluation.''

\subsection{Submission Path}

Qiushi was run through its Web~UI, so the original agent sessions were not natively instrumented by Inspect. The scientific outputs (report, code, trace, and artifacts) were exported from each Qiushi session without content edits and scored from the command line using the official AstaBench E2E rubric scorer (\texttt{astabench score}). The 40 single-sample scorer logs were then merged into the official E2E-Bench-Hard test task shape.

Each sample in the submitted \texttt{.eval} archive contains the original scorer events plus one aggregate, non-scorer \texttt{ModelEvent} representing the actual Qiushi UI input/output token totals. These totals were transcribed from corresponding Qiushi Overview screenshots and represent approximate values (token counts shown with K/M units are rounded). This is an aggregate usage record, not a claim that Inspect captured each native DeepSeek API request.

\subsection{Official Cost and Resource Usage}

\begin{table}[htbp]
\centering
\caption{Resource usage summary across all 40 E2E-Bench-Hard test runs.}
\label{tab:resources}
\begin{tabularx}{\textwidth}{L{5.5cm}Y}
\toprule
\textbf{Metric} & \textbf{Value} \\
\midrule
Official leaderboard cost per task & \$15.209 \\
Total full LLM tokens (approx.) & 1.376 billion \\
Mean tokens per sample & 34.4 million \\
Total wall-clock time & 386.25 hours \\
Mean duration per sample & 9.66 hours \\
Mean recorded workflow steps per sample & 27.65 \\
Mean turns per sample & 212.43 \\
Mean tool calls per sample & 223.25 \\
Mean LLM calls per sample & 505.12 \\
Samples completed normally & 37 (completed\_verified) \\
Samples reaching step limit & 3 (step\_limit\_reached) \\
\bottomrule
\end{tabularx}
\end{table}

\Cref{tab:resources} summarizes the official benchmark cost and recorded resource usage. Monetary cost is reported using the official AstaBench leaderboard value of \$15.209 per task.\cite{astabenchleaderboard} Token counts, duration, and workflow counts are drawn from the run telemetry.

The UI mode distribution across the 40 runs was: Coding (16 samples), Discovery (12), Submission (11), and Report (1). The scientific research strategist setting was Semi Active for 23 samples and Non Active for 17.

\subsection{Inspect Evaluation Archive Structure}

The submitted \texttt{.eval} file is a 112~MB ZIP-compressed Inspect log containing 45~entries:

\begin{itemize}
\item \texttt{\_journal/start.json}---the Inspect session startup log recording evaluation start time and configuration parameters.
\item 40~\texttt{samples/idea\_*\_epoch\_1.json}---each corresponding to one E2E-Bench-Hard test sample's complete evaluation record. Each sample JSON contains:
  \begin{itemize}
  \item Task definition (name, description, hypothesis, variables, contrasts, measurements)
  \item Input text (complete task specification)
  \item Output structure (report text, code file list, artifact file list)
  \item Scorer events (per-rubric-item judgments and explanation text)
  \item Usage statistics (token counts and event counts)
  \end{itemize}
\item \texttt{summaries.json}---score summaries for all 40~samples.
\item \texttt{reductions.json}---score aggregation results (mean, standard error).
\item \qpath{header.json}---evaluation metadata: task \qpath{astabench/e2e_discovery_hard_test}, model \qpath{deepseek/deepseek-v4pro-preview}, and scorer \qpath{score_rubric}.
\end{itemize}

The \texttt{model} field in \texttt{header.json} records the Inspect/task metadata label \texttt{deepseek/deepseek-v4pro-preview}, corresponding to the backend configured in the Qiushi Engine Web UI. The offline replay-mode run objects use \texttt{mockllm/model}; these are different roles.

Each sample's scorer explanation text is the raw data source for the rubric-item analysis in this report. The scorer produces text in the format: ``\texttt{[MET] Dataset Implementation: The code contains a data loader...}'' or ``\texttt{[NOT MET] Statistical Testing: No evidence of multiple runs...}''. Script~4 (\texttt{extract\_all\_rubric\_process\_patterns.py}) parses these texts via regular expressions to extract rubric item names and binary judgments.

\subsection{UI Run Mode Analysis}

The Qiushi Engine Web UI supports multiple run modes, and the submitter selected the most appropriate mode for each task. The mode distribution across the 40~runs was: Coding (16~samples, 40\%), Discovery (12, 30\%), Submission (11, 27.5\%), Report (1, 2.5\%).

\begin{table}[htbp]
\centering
\caption{Score distribution by UI run mode. The four modes show minimal score differences, indicating that mode selection had limited impact on final scores.}
\label{tab:ui-modes}
\small
\begin{tabularx}{\textwidth}{L{2.5cm}ccccY}
\toprule
\textbf{UI Mode} & \textbf{n} & \textbf{Mean} & \textbf{Range} & \textbf{Median} & \textbf{Description} \\
\midrule
Coding & 16 & 0.810 & 0.54--1.00 & 0.83 & Code-implementation-focused mode \\
Discovery & 12 & 0.809 & 0.50--1.00 & 0.83 & Scientific-discovery-focused mode \\
Submission & 11 & 0.822 & 0.67--1.00 & 0.83 & Final-submission-focused mode \\
Report & 1 & 0.941 & 0.94 & 0.94 & Report-writing-focused mode \\
\bottomrule
\end{tabularx}
\end{table}

The observed mode means are close for Coding, Discovery, and Submission; Report has only $n{=}1$. Because mode selection was not randomized and is confounded with task choice, these descriptive values do not identify an effect of UI mode on rubric coverage.

\subsection{Evidence Sources and Supported Statements}
\begin{table}[htbp]\centering\small
\caption{Evidence sources used in this report and the statements each source can support.}
\label{tab:evidence-sources}
\begin{tabularx}{\textwidth}{L{3.4cm}L{4.7cm}Y}\toprule
\textbf{Source} & \textbf{Supports} & \textbf{Does not establish} \\ \midrule
\texttt{summary\_stats.json} & Full-precision macro score and standard error & Consistent with official displayed score \\
40 sample JSONs & Per-item judgments and scorer explanations & Human-expert correctness of the judge \\
\qpath{test_model_configuration.json} & Submitted model-configuration identifiers & Every native API request \\
UI telemetry & Approximate tokens and duration & Standardized cross-agent cost efficiency \\
Meta-Trace files & Recorded workflow states and transitions & Causal contribution of any workflow component \\
SHA-256 inventory & File identity and integrity & Scientific correctness of file contents \\
\bottomrule\end{tabularx}\end{table}

\section{Aggregate Results}
\label{sec:aggregate}

\subsection{Overall Score}

The locally recomputed AstaBench task score for Qiushi Engine v0.8 on E2E-Bench-Hard is:
\begin{equation}
\text{Score} = 0.8159 \pm 0.0187 \quad (\text{mean} \pm \text{standard error}, \; n = 40)
\label{eq:score}
\end{equation}

This score was computed by the pinned official \texttt{astabench score} command. Its mean rounds to the 0.816 value now displayed on the official leaderboard. The full-precision result is equivalently expressed as $81.59 \pm 1.87$ on the percentage scale. The median sample score is 0.833, the minimum is 0.500, and the maximum is 1.000, with a sample standard deviation of 0.119.

Beyond the mean score, full-task completion is a stricter complementary measure. Four of the 40 tasks satisfied every required rubric item, giving Qiushi Engine a $4/40=10\%$ perfect-completion rate. The AstaBench paper reports an approximately 3\% maximum full-task completion rate for its official agents on E2E-Bench-Hard (see \S\ref{sec:benchmark}); against that published reference, Qiushi is 7 percentage points higher, or about 3.3 times the rate. This task-level perfect-completion statistic is distinct from the 82.1\% micro-average fulfillment rate across individual rubric items.

Of the 40~samples, 37 reached \texttt{completed\_verified} status (the Qiushi session concluded normally after the required completion checks) and 3 reached \texttt{step\_limit\_reached} (the session exhausted its step budget before the completion protocol finished). Notably, one of the three step-limited samples (\texttt{idea\_5}) still achieved a perfect score of~1.0, demonstrating that reaching the step limit does not necessarily imply incomplete research output---it may simply mean that the verification protocol was interrupted after the research work was already complete.

\subsection{Score Distribution and Bands}

\begin{figure}[htbp]
\centering
\includegraphics[width=0.78\textwidth]{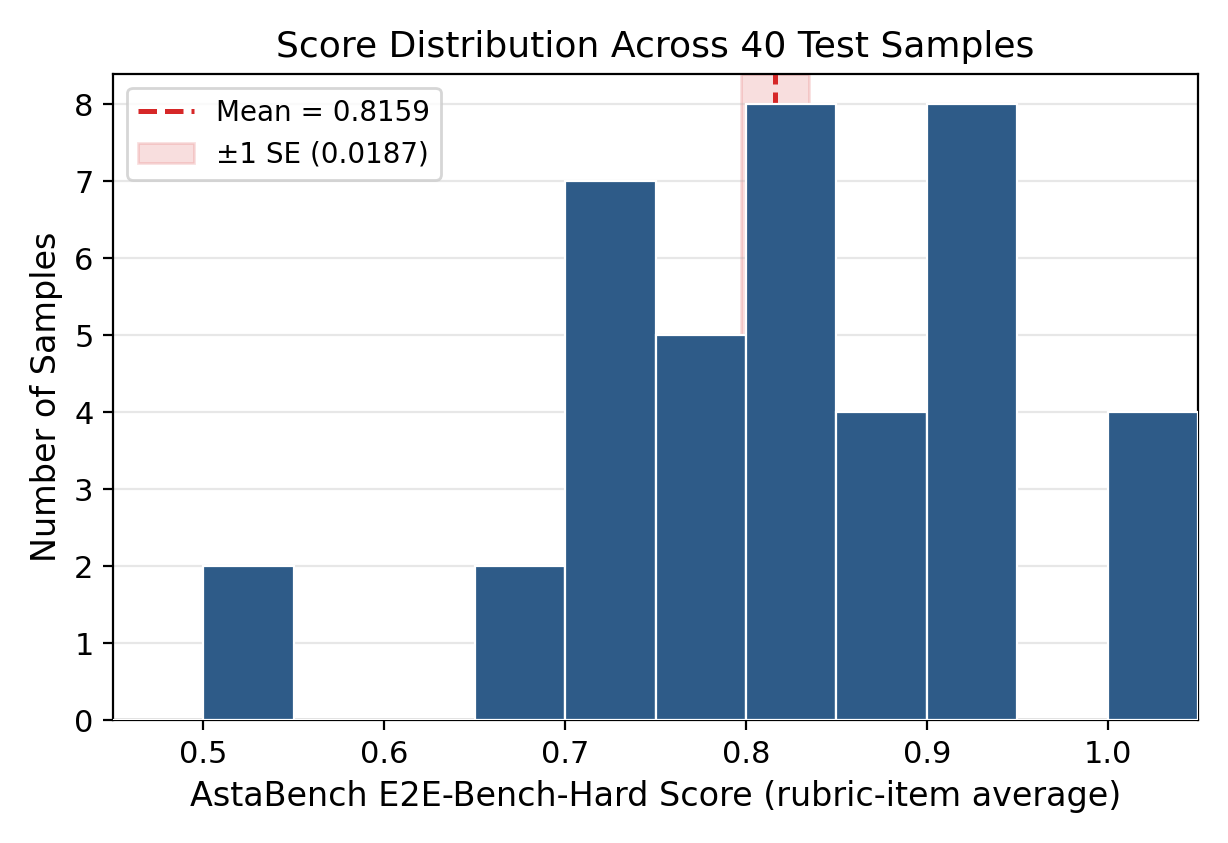}
\caption{Distribution of rubric-item average scores across the 40 E2E-Bench-Hard test samples. The dashed line marks the mean score (0.8159); the shaded band shows $\pm 1$ standard error. Scores range from 0.50 to 1.00, with a median of 0.833.}
\label{fig:score-dist}
\end{figure}

\Cref{fig:score-dist} shows the score distribution. We classify the 40 samples into five performance bands:

\begin{itemize}
\item \textbf{All items met} (score = 1.0): 4 samples---\texttt{idea\_5}, \texttt{idea\_22}, \texttt{idea\_29}, \texttt{idea\_42}.
\item \textbf{Near-complete} ($0.90 \leq$ score $< 1.0$): 8 samples.
\item \textbf{Strong} ($0.80 \leq$ score $< 0.90$): 12 samples.
\item \textbf{Substantial but incomplete} ($0.70 \leq$ score $< 0.80$): 12 samples.
\item \textbf{Constrained} (score $< 0.70$): 4 samples---\texttt{idea\_12}, \texttt{idea\_16}, \texttt{idea\_10}, \texttt{idea\_14}.
\end{itemize}

\begin{figure}[htbp]
\centering
\includegraphics[width=0.68\textwidth]{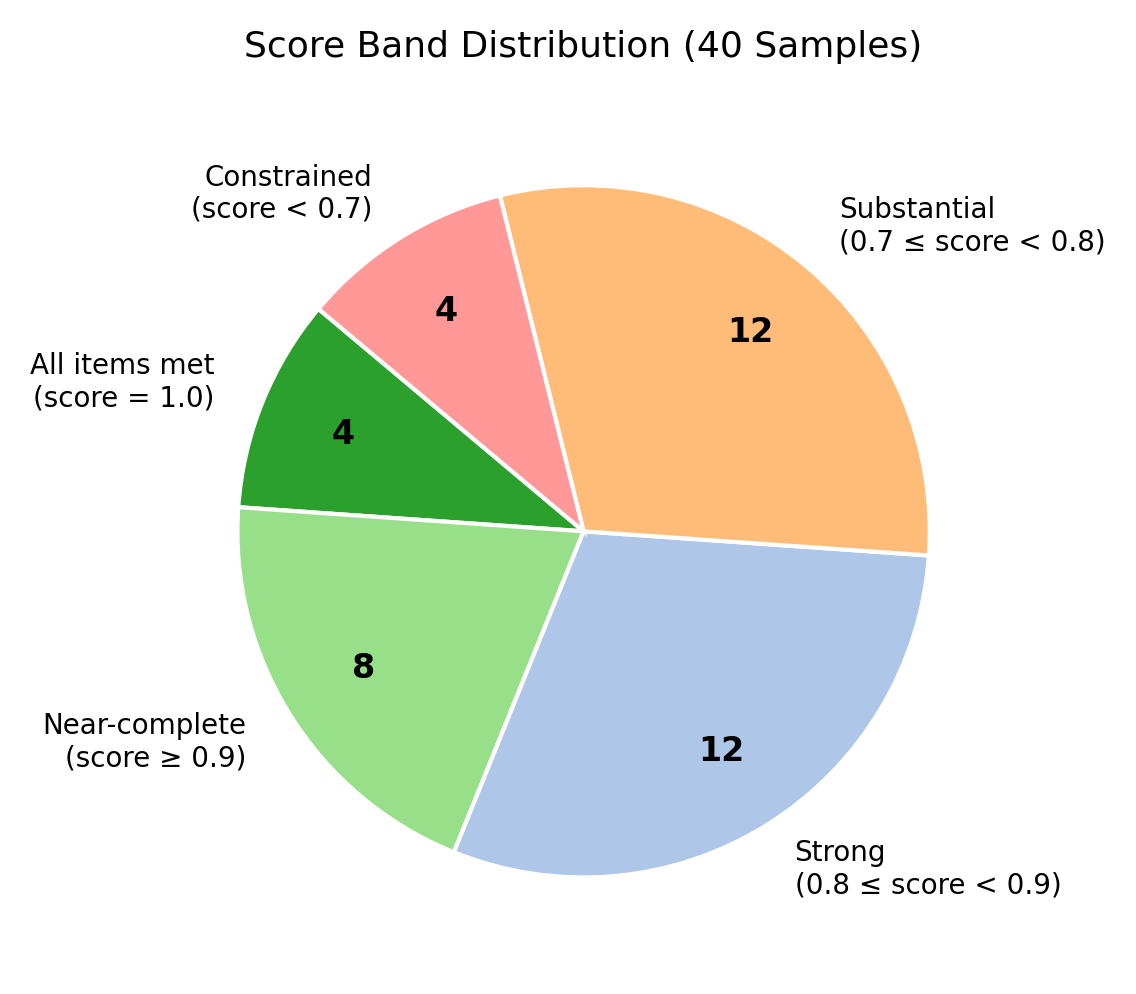}
\caption{Score band distribution. The majority of samples (24/40, 60\%) achieved strong or better rubric coverage ($\geq 0.80$).}
\label{fig:score-bands}
\end{figure}

\begin{figure}[htbp]
\centering
\includegraphics[width=\textwidth]{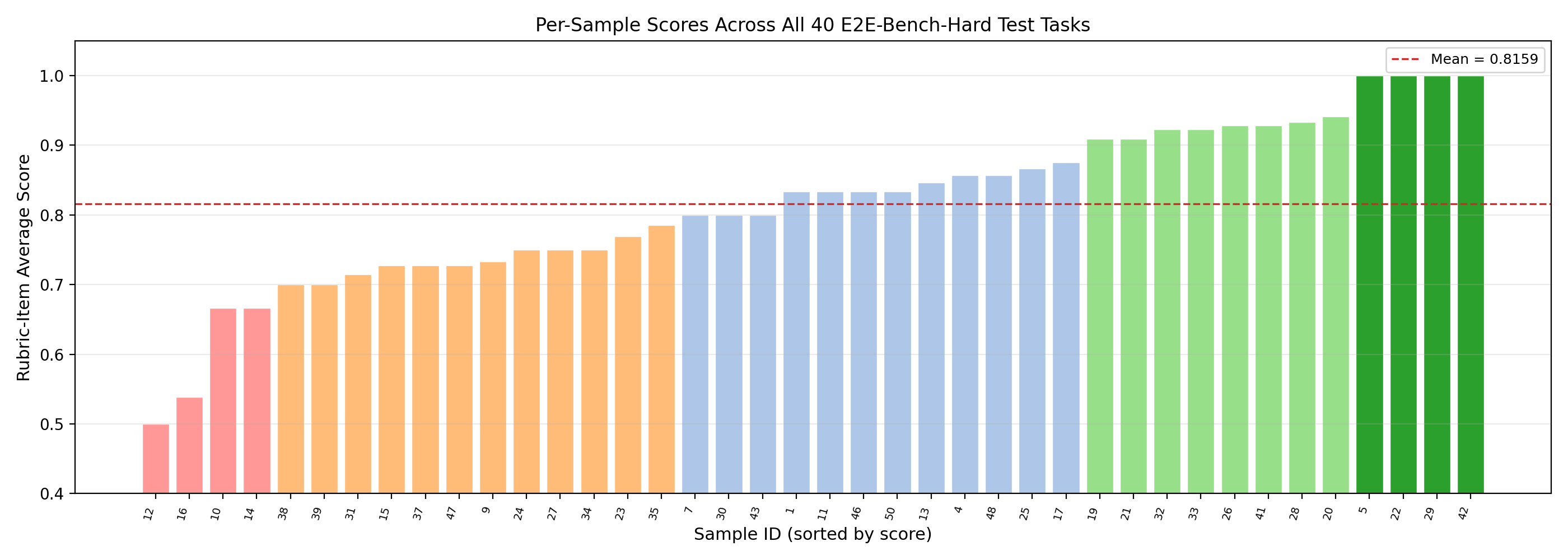}
\caption{Per-sample rubric-item average scores across all 40 test tasks, sorted in ascending order. Color coding corresponds to the five score bands. The dashed line marks the overall mean.}
\label{fig:per-sample-scores}
\end{figure}

The distribution is left-skewed: 60\% of samples cluster at or above 0.80, with a small tail of constrained runs (\Cref{fig:score-bands}). \Cref{fig:per-sample-scores} shows the individual sample scores in sorted order, providing a complete view of the score landscape.

\subsection{Score--Resource Relationships}

A natural question is whether resource expenditure covaries with score. In this 40-sample archive the Pearson correlations are close to zero: $r=0.001$ for tokens and $r=-0.010$ for steps. These descriptive correlations show no observed linear association; the non-random tasks and resource allocation do not support causal interpretation.

\begin{figure}[htbp]
\centering
\includegraphics[width=\textwidth]{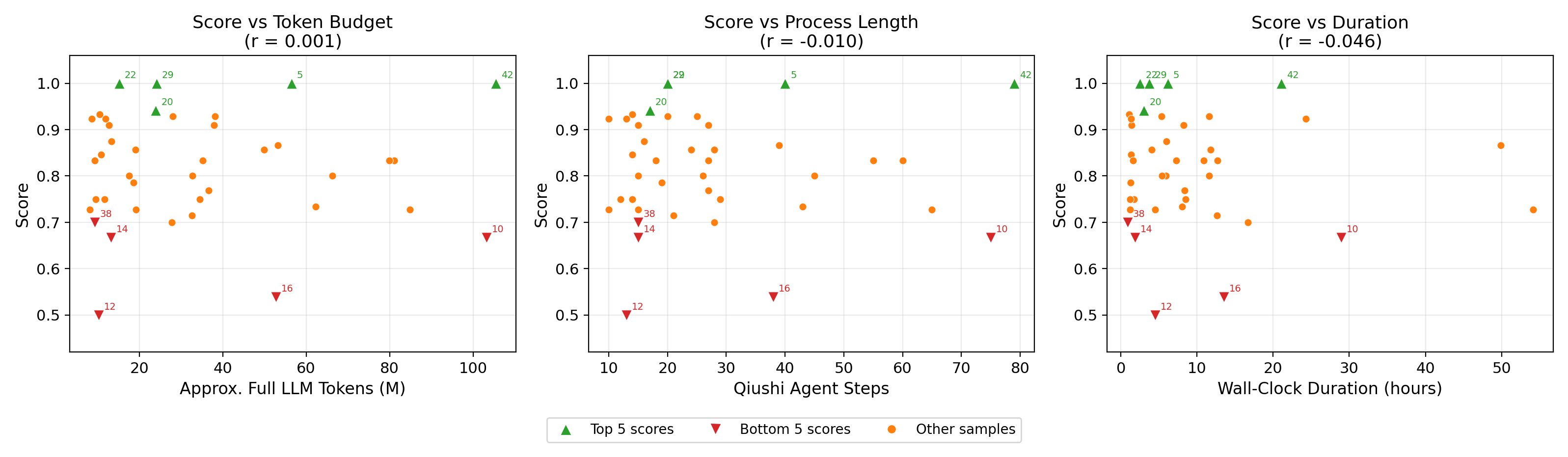}
\caption{Score vs.\ resource expenditure across three dimensions: token budget, process length (agent steps), and wall-clock duration. Green triangles mark the five highest-scoring samples; red triangles mark the five lowest-scoring. Correlations are near zero in all cases, indicating that resource volume did not predict score within this evaluation.}
\label{fig:score-workload}
\end{figure}

The labeled samples in \Cref{fig:score-workload} illustrate this vividly: \texttt{idea\_22} scored 1.0 with only 15.2M tokens and 20 steps, while \texttt{idea\_10} scored 0.667 despite consuming 103.2M tokens and 75 steps. These examples show that resource volume alone is insufficient to explain score variation in this archive.

\subsection{Resource Usage Distributions}

\begin{figure}[htbp]
\centering
\includegraphics[width=\textwidth]{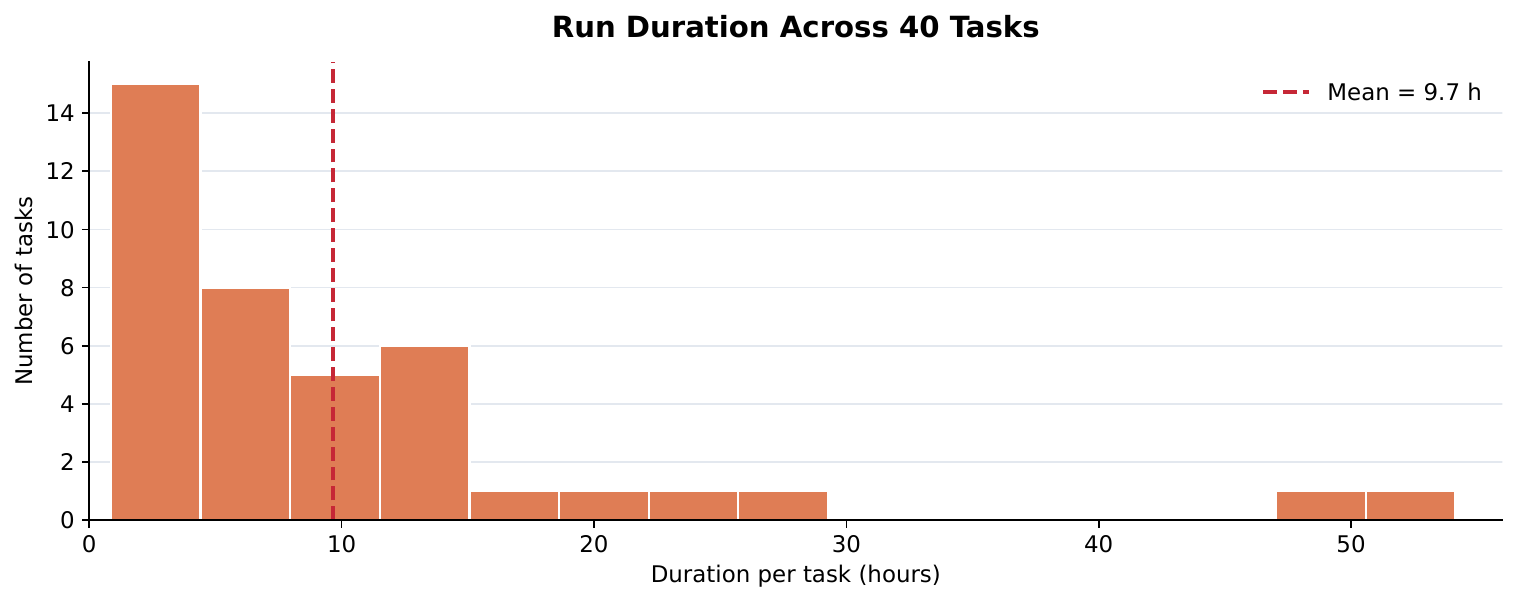}
\caption{Duration distribution across the 40 samples. The right-skewed distribution contains a small number of long-running tasks. The dashed line marks the mean duration.}
\label{fig:duration-distribution}
\end{figure}

\Cref{fig:duration-distribution} shows the duration distribution. The median duration per sample is 5.9 hours (mean 9.7 hours). Most samples complete within a shorter period, while a few samples (\texttt{idea\_10} at 28.9 hours, \texttt{idea\_37} at 54.1 hours, \texttt{idea\_25} at 49.9 hours) take substantially longer, often due to extended repair cycles.

\subsection{Score Variation Across Task Families}

The 40~test tasks span diverse AI/NLP research areas, and their rubric coverage varies systematically by task family (\Cref{tab:family-analysis}). This variation is important because it reveals whether Qiushi's performance depends on the research domain or is broadly consistent.

\begin{table}[htbp]
\centering
\caption{Per-task-family score analysis. Families are sorted by sample count. ``Zero/Total'' shows zero-scored items out of total scored items across all samples in the family.}
\label{tab:family-analysis}
\small
\begin{tabularx}{\textwidth}{L{4.2cm}cccL{1.2cm}L{1.5cm}Y}
\toprule
\textbf{Task Family} & \textbf{n} & \textbf{Mean} & \textbf{Range} & \textbf{Zero/ Total} & \textbf{Mean Steps} & \textbf{Dominant Zero-Score Pattern} \\
\midrule
Reasoning / symbolic problem solving & 10 & 0.840 & 0.70--1.00 & 19/121 & 29.3 & Statistical support; repeated runs \\
Other NLP/ML research & 8 & 0.826 & 0.73--0.93 & 18/107 & 21.0 & Mixed: metrics, repeated runs \\
Classification / robustness & 7 & 0.828 & 0.50--1.00 & 15/91 & 22.0 & Model requirements; statistics \\
Dialogue / retrieval QA & 5 & 0.752 & 0.67--0.83 & 16/63 & 49.6 & Core components; metrics; statistics \\
Summarization / factual consistency & 5 & 0.819 & 0.67--1.00 & 9/59 & 27.8 & Ablation; core module \\
Info.~extraction / clinical NLP & 3 & 0.684 & 0.54--0.80 & 13/42 & 30.3 & External model dependencies \\
Agents / decision-making & 1 & 1.000 & 1.00 & 0/10 & 20.0 & None \\
Security / privacy & 1 & 0.929 & 0.93 & 1/14 & 25.0 & Ablation (poisoning rates) \\
\bottomrule
\end{tabularx}
\end{table}

Several patterns emerge from \Cref{tab:family-analysis}:

\textbf{Information extraction and clinical NLP} has the lowest family mean (0.684) with only 3~samples. This family includes \texttt{idea\_16} (Anticipatory Ensemble for Medication Extraction, score 0.538), which suffered the most severe cascading dependency failure in the archive. The clinical NLP tasks frequently require specific medical models (GPT-3.5, LLaMA-2) that were unavailable, making this family disproportionately affected by external dependency failures.

\textbf{Dialogue and retrieval QA} has the second-lowest mean (0.752) despite the highest mean step count (49.6). This family includes resource-intensive samples like \texttt{idea\_1} (55~steps) and \texttt{idea\_10} (75~steps), where complex multi-component systems required extensive implementation and repair effort, leaving less budget for empirical breadth. The high zero-item ratio (16/63 = 25.4\%) reflects this pattern.

\textbf{Reasoning and symbolic problem solving} is the largest family (10~samples) and scores well (mean 0.840), with three of the four perfect-scoring samples belonging to this family. The zero-scored items here are predominantly statistical support and repeated runs---the least severe category of score loss.

\textbf{Classification and robustness} shows the widest score range (0.50--1.00), spanning from \texttt{idea\_12} (TAPP, the lowest-scoring sample overall) to \texttt{idea\_5} (Dynamic Commonsense Integration, a perfect score). This within-family variation indicates that the task-specific requirements, not the research domain alone, determine the outcome.

The overall pattern suggests that Qiushi's performance is broadly consistent across task families, with the primary variation driven by task-specific external dependencies and complexity rather than by research domain. The one exception is information extraction/clinical NLP, where specific model requirements create a structural disadvantage.

\section{Process Analysis: What Produced High Scores and What Caused Losses}
\label{sec:process}

The central analytical contribution of this report is connecting rubric-level outcomes to Qiushi's recorded research workflow. Since each rubric item corresponds to a specific research action, the archive makes it possible to trace the evidence chain from planning, implementation, execution, and verification to the final score.

\subsection{Rubric-Item Landscape}

Across the 40~test samples, the AstaBench scorer evaluated 507~rubric items in total. Of these, 416 were scored as met and 91 as zero (micro-average fulfillment $416/507 = 82.1\%$, which is close to but distinct from the official macro-average task score of $81.59\%$). The rubric items span a wide range of research actions: dataset loading and preparation, method implementation, baseline construction, metric computation, statistical testing, ablation studies, error analysis, visualization, documentation, and more.

\begin{figure}[htbp]
\centering
\includegraphics[width=\textwidth]{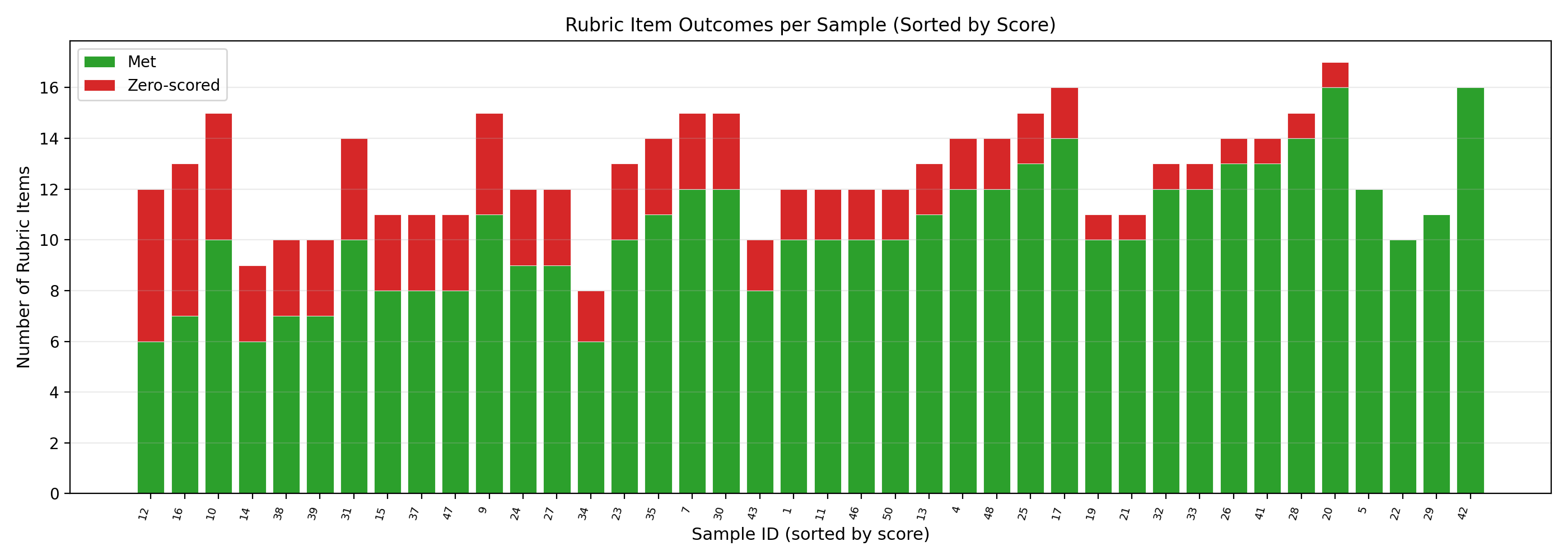}
\caption{Stacked bar chart showing the number of met (green) and zero-scored (red) rubric items per sample, sorted by score. Samples on the right have all items met; samples on the left have the most zero-scored items. The total bar height reflects the number of scored rubric items, which varies by task.}
\label{fig:rubric-stacked}
\end{figure}

\Cref{fig:rubric-stacked} shows the rubric item outcomes per sample. The variation in total bar height reflects the different number of scored rubric items per task (range: 8--17). The pattern is clear: higher-scoring samples have fewer zero-scored items, but even some of the lower-scoring samples (e.g., \texttt{idea\_10} with 10/15 items met) satisfied the majority of rubric items, losing points on specific identifiable gaps rather than failing broadly.

\subsection{Recurring Zero-Scored Dimensions}

\begin{figure}[htbp]
\centering
\includegraphics[width=0.88\textwidth]{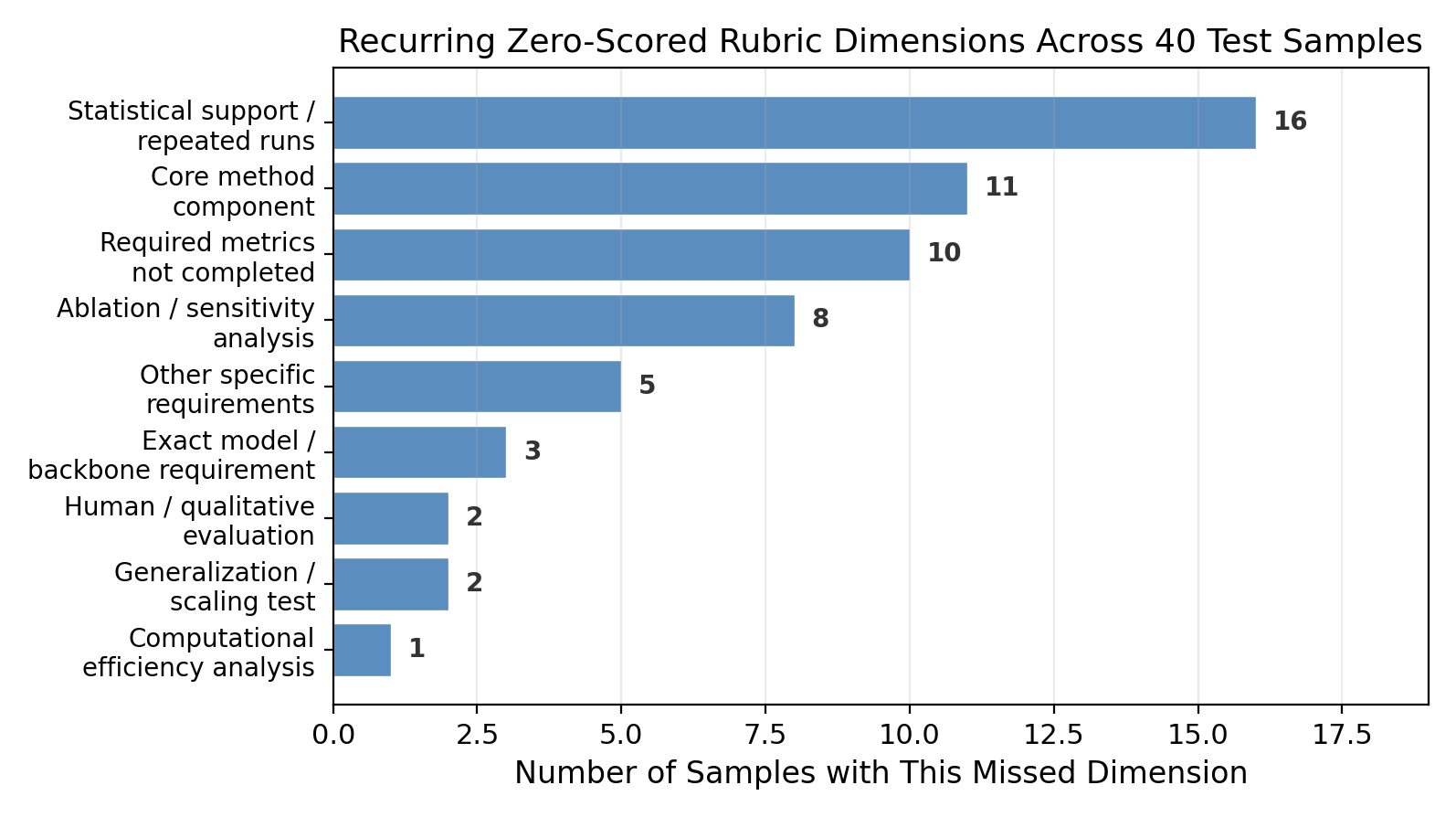}
\caption{Recurring categories of zero-scored rubric dimensions across 40 samples. ``Statistical support / repeated runs'' was the most commonly missed category (16 samples), followed by ``core method component'' requirements (11 samples) and ``required metrics not completed'' (10 samples).}
\label{fig:zero-dims}
\end{figure}

\Cref{fig:zero-dims} shows the distribution of zero-scored items by category. The pattern is not uniform---certain types of research actions were systematically more likely to be missed:

\textbf{Statistical support and repeated runs} (16 samples). Many tasks require multiple independent experimental runs with statistical significance testing (e.g., paired t-tests, bootstrap resampling, or computation of confidence intervals). These items demand that the agent not only implement the experiment once but run it multiple times with different seeds or conditions and then compute inter-run statistics. When the step budget is limited or the primary implementation consumes most of the available compute, repeated runs and their statistical analysis are often the first items to be sacrificed.

\textbf{Core method components requiring external resources} (11 samples). Some rubric items require implementing specific components that depend on external models, datasets, or APIs. For example, a ``Neural Selector'' baseline in a retrieval task, or a ``Knowledge Graph Integration'' module in a verification framework. When the required resource is unavailable in the execution environment or the component's implementation is too complex to complete within the step budget, the item scores zero.

\textbf{Explicit metric completion} (10 samples). Certain rubric items require specific metrics (e.g., response accuracy on a particular split, F1 with a specific tokenization, or a named evaluation like BERTScore) that were either not computed, produced values the scorer could not confirm in the output artifacts, or were computed but not properly reported in the final deliverables.

\textbf{Ablation and sensitivity analyses} (8 samples). Ablation studies require multiple experimental conditions beyond the core comparison---removing or varying individual components to isolate their contributions. Like repeated runs, they are typically scheduled late in the research process and may be truncated by step or time limits.

\textbf{Exact model/backbone requirements} (3 samples). Some tasks specify particular model architectures (e.g., GPT-3.5, LLaMA-2, GPT-4) that must be used as components of the experiment. When these models cannot be accessed through the available API or execution environment, all downstream items dependent on those models score zero.

\textbf{Other categories} include human or qualitative evaluation (2 samples), generalization or cross-domain testing (2), and computational efficiency analysis (1).

\subsection{Strength Pattern: Core Implementation and Artifact Production}

Qiushi consistently scored well on rubric items involving: dataset acquisition and preparation, core method implementation, baseline system construction, integration mechanisms, primary metric computation, and report/code/artifact bundle production. These are the backbone of each research task---the items that require designing, coding, running, and packaging a working experiment.

This pattern is consistent with an artifact-centered workflow in which method construction, execution, repair, and final consistency checking are connected. The Express phase packages the report, code, trace, and artifacts into the deliverable structure expected by the AstaBench scorer. This workflow maps naturally onto the rubric's requirement that claims be supported across report, code, and artifact evidence; the association is descriptive and does not isolate a causal architectural effect.

\subsection{Process Characteristics Across Score Bands}

The Meta-Trace state summaries show that the constrained and substantial-but-incomplete score bands contain more Repair and Verify activity on average. This count pattern is consistent with more frequent execution difficulties in the inspected cases, but the counts alone do not identify the cause of score loss.

Interestingly, the all-items-met band does not show the lowest resource usage. Some perfect-scoring samples (notably \texttt{idea\_42} with 79 steps) required extensive repair cycles to achieve complete rubric coverage, while others (\texttt{idea\_22} with 20 steps) completed efficiently. This further supports the finding that score is determined by the completeness and correctness of research actions, not by resource volume.

\subsection{Case Studies: Perfect-Scoring Samples}

\begin{table}[htbp]
\centering
\caption{Four perfect-scoring samples with rubric details and process characteristics. All scored rubric items were met in each case.}
\label{tab:perfect-samples}
\small
\begin{tabularx}{\textwidth}{L{1.4cm}L{3.2cm}cccccY}
\toprule
\textbf{ID} & \textbf{Task Name} & \textbf{Score} & \textbf{Met} & \textbf{Tot} & \textbf{Steps} & \textbf{Tok (M)} & \textbf{Key Process Feature} \\
\midrule
idea\_5 & Dynamic Commonsense Integration & 1.00 & 12 & 12 & 40 & 56.5 & CUDA determinism fix; paired t-tests; seed sensitivity analysis \\
idea\_22 & Memory-Aug.\ Decentral.\ Decision & 1.00 & 10 & 10 & 20 & 15.2 & SHA-256 file verification; causally distinct conditions \\
idea\_29 & Memory-Enhanced Tree Reasoning & 1.00 & 11 & 11 & 20 & 24.1 & Final-deliverable re-verification; artifact inspection \\
idea\_42 & FactCC-CaPE Integration & 1.00 & 16 & 16 & 79 & 105.4 & Dual-dataset eval; matched-alpha controls; 79-step repair cycle \\
\bottomrule
\end{tabularx}
\end{table}

The perfect-scoring samples (\Cref{tab:perfect-samples}) share several process characteristics visible in their Meta-Trace records:

\textbf{idea\_5} (Dynamic Commonsense Integration, score 1.0, 40 steps). This task required integrating ConceptNet and ATOMIC knowledge sources with a context-aware emotional graph attention mechanism. The final Meta-Trace records that all 26 experimental checks were passed, including a determinism fix for CUDA non-reproducibility, seed-sensitivity analysis across multiple runs, and paired t-tests. The agent identified that the ``Dynamic'' model showed extreme seed sensitivity (F1 ranging from 0.000 to 0.066 across seeds), reported this as a genuine phenomenon rather than an experimental artifact, and included the NoKnowledge baseline comparison. Despite reaching the step limit (40 steps), all scored rubric items were satisfied.

\textbf{idea\_22} (Memory-Augmented Decentralized Decision-Making, score 1.0, 20 steps). This task asked the agent to integrate memory-augmented planning with GPT-based decision engines in the Overcooked-AI environment. The Preceptor's final verification confirmed SHA-256 hashes of all 10 code files and verified that the three experimental conditions (GPT-only, memory-only rule-based, and memory-augmented GPT) were causally distinct in the runner code. The efficient 20-step completion suggests a well-matched task complexity and execution capability.

\textbf{idea\_42} (FactCC-CaPE Integration, score 1.0, 79 steps). This task recorded 79 steps, the highest step count in the archive. The agent needed to implement FactCC-based factual consistency scoring, CaPE fine-tuning, expert/anti-expert model training, and evaluate on both XSUM and CNN/DM datasets. The matched-alpha control experiment (ensuring that the integrated model's advantage was not simply due to a better alpha parameter) added complexity. The Meta-Trace shows 25 Repair states---the highest in the archive---indicating extensive debugging and iteration before the full rubric was satisfied.

\subsection{Case Studies: Medium-Scoring Samples}

The medium-scoring samples (0.73--0.87) are particularly informative because they reveal partial success patterns---tasks where the core research was substantially completed but specific dimensions fell short.

\textbf{idea\_1} (Hierarchical Hybrid QA Integration, score 0.833, 10/12 items met, 55 steps). This task required implementing a hierarchical table processing module for multi-hop QA over the MultiHiertt dataset. The two zero-scored items were ``Hybrid Evidence Utilization Algorithm'' and ``Multiple Experimental Runs.'' The agent successfully implemented the dataset, processing modules, baselines, metrics, and statistical comparison, but the specific hybrid evidence algorithm and repeated independent runs were not completed. With 55 steps and 81.1M tokens, this was a resource-intensive run, suggesting the task's complexity exceeded what could be fully covered within the budget.

\textbf{idea\_37} (score 0.727, 8/11 items met, 65 steps, 84.9M tokens, 54.1 hours). The zero-scored items were ``F1 Score Calculation,'' ``Reframing Technique Documentation,'' and ``Coherence Score Thresholds.'' Its observable Meta-Trace contains 48 Process, 5 Repair, 4 End, 4 Evaluate, 3 Plan, and 1 Review states, showing that high resource use did not substitute for satisfying the three specific requirements.

\textbf{idea\_25} (Integrated Model Editing for Consistency, score 0.867, 13/15 items met, 39 steps, 49.9 hours). The second-longest run by duration. Missed items were ``Multiple Independent Runs'' and ``Meta-Analysis''---both statistical support requirements. The core MEMIT implementation, RippleEdits benchmark integration, and all measurement items were met.

\textbf{idea\_7} (Cross-Document Structured Extraction, score 0.800, 12/15 items met, 45 steps). The three missed items were dataset acquisition, statistical significance testing, and cross-validation. This illustrates a cascade pattern: if a specific dataset cannot be acquired, downstream items dependent on that data (statistical testing, cross-validation) also fail.

\textbf{idea\_50} (Augmented Graph-Based Schema Encoding, score 0.833, 10/12 items met, 60 steps, step\_limit\_reached). One of the three step-limited samples. Missed items were the SGD-X dataset implementation and repeated runs. Despite reaching the step limit, 83\% of rubric items were satisfied, indicating substantial research progress before termination.

The near-complete samples ($\geq 0.90$) are instructive because they show where the system came within one or two rubric items of a perfect score and what prevented full coverage.

\textbf{idea\_20} (Syntactic Trigger Inspection, score 0.941, 16/17 items met, 17 steps, 24.0M tokens). This task investigated backdoor attack techniques using syntactic triggers and model inspection. The single zero-scored item was ``Varying Poisoning Rates''---an experimental condition the agent did not implement. The Preceptor's final reasoning noted that the experiment produced a legitimate negative result: the backdoor technique was too weak to demonstrate the hypothesized effect, and the report stated this honestly rather than overclaiming. This case illustrates that near-perfect scores can coexist with scientifically honest negative findings. The Meta-Trace shows only 6~Process and 3~Repair states, indicating smooth implementation.

\textbf{idea\_28} (Contrastive Bayesian Hybrid Learning, score 0.933, 14/15 items met, 14 steps, 10.5M tokens). This active learning task scored well across implementation, baselines, metrics, and statistical testing. The single zero-scored item was ``Low-Resource Dataset Implementation.'' Completed in only 14~steps and 1.1~hours, this was one of the most efficient runs in the archive. The parsed state counts for this row are unavailable in Appendix~\ref{app:process-details}, so the evidence here is the score/resource record rather than a state-pattern reconstruction.

\textbf{idea\_26} (Integrated Contrastive Learning for Factual Extraction, score 0.929, 13/14 items met, 20 steps). The single missed item was ``Multiple Experimental Runs.'' Core implementation, baselines, all metrics, and statistical significance testing were completed. This case typifies the ``only missing repeated runs'' pattern---the most common single-item failure across the 40~samples.

\textbf{idea\_32} (Dynamic Streaming Evaluation, score 0.923, 12/13 items met) was especially short, with 10~steps, 8.6M tokens, and 1.4~hours. \textbf{idea\_33} (Adversarial Contextual Embeddings, score 0.923, 12/13 items met) also used few steps---13~steps and 11.9M tokens---but had a longer wall-clock duration of 24.3~hours. For idea\_33, the single missed item was ``Coded Hate Speech Detection Analysis''; for idea\_32, it was ``Computational Efficiency Measurement.'' These samples show that short, high-scoring runs can still miss a task-specific analysis or efficiency metric.

\textbf{idea\_41} (FastText Membership Inference, score 0.929, 13/14 items met, 25 steps). The single missed item was ``Multiple Dataset Validation.'' Core FastText embedding extraction, membership inference attack implementation, privacy metrics, and statistical analysis were all complete. The pattern is consistent: when one specific experimental extension (additional dataset, repeated runs, or ablation condition) is not completed, the resulting score loss is small (one item out of 14) but prevents a perfect score.

The near-complete samples share a common profile: core implementation is complete, primary metrics are computed, basic significance analysis is performed, and the only gap is one specific experimental condition, dataset variant, or repeated-run requirement. Within these samples, core implementation items had high observed coverage; remaining gaps often concerned an additional condition, dataset, or repeated run.

\subsection{Case Studies: Constrained Samples}

\begin{table}[htbp]
\centering
\caption{The four constrained samples (score $< 0.70$) with zero-scored dimensions and process details.}
\label{tab:constrained-samples}
\small
\begin{tabularx}{\textwidth}{L{1.4cm}L{2.5cm}cL{0.9cm}ccL{4.5cm}}
\toprule
\textbf{ID} & \textbf{Task Name} & \textbf{Score} & \textbf{Met/ Tot} & \textbf{Steps} & \textbf{Tok (M)} & \textbf{Zero-Scored Items} \\
\midrule
idea\_12 & TAPP with Explicit Demos & 0.50 & 6/12 & 13 & 10.2 & LLM Selection; Accuracy; F1; Statistical Testing; Replication; Demo Selection Strategy \\
idea\_16 & Anticipatory Ensemble Med.\ Extract. & 0.54 & 7/13 & 38 & 52.8 & GPT-3.5 Integration; LLaMA-2 Integration; Repeated Runs; Meta-Analysis; Stat.\ Testing; Model Size Comparison \\
idea\_10 & Contextual Feedback Integration & 0.67 & 10/15 & 75 & 103.2 & Response Accuracy; Repeated Runs; Ablation; Trigger Conditions; Qualitative Analysis \\
idea\_14 & Adaptive FactCC Decoding & 0.67 & 6/9 & 15 & 13.2 & FactCC Eval Module; Ablation; Abstractiveness \\
\bottomrule
\end{tabularx}
\end{table}

The constrained samples (\Cref{tab:constrained-samples}) reveal distinct failure modes:

\textbf{idea\_12} (TAPP with Explicit Demonstrations, score 0.500, 6/12 items met). This task asked the agent to enhance LLM text classification using TAPP with explicit answer demonstrations. The agent implemented TAPP, baseline configurations, both datasets, cross-task demonstrations, and prompt format documentation. However, it did not receive credit for LLM model selection (the task required a specific large model), accuracy and F1 measurement, statistical testing, experimental replication, or demonstration selection strategy. With only 13 steps and 10.2M tokens, this was one of the shortest runs, suggesting that the session may have concluded prematurely---possibly interpreting the task too narrowly or failing to recognize that measurement and statistical requirements were not yet satisfied.

\textbf{idea\_16} (Anticipatory Ensemble for Medication Extraction, score 0.538, 7/13 items met). This is a classic cascading-dependency failure. The task required integrating both GPT-3.5 and LLaMA-2 as model backends for medication extraction. When these specific models could not be accessed in the execution environment, six downstream rubric items---model integration, repeated runs across models, meta-analysis, statistical testing, and model-size comparison---all scored zero. The core prompting framework, ensemble voting, and precision/recall computation were implemented and scored well, demonstrating that the research design was sound but the external dependency chain was broken.

\textbf{idea\_10} (Contextual Feedback Integration, score 0.667, 10/15 items met, 75 steps, 103.2M tokens). This was the second-highest resource consumer in the archive. The Meta-Trace shows 22 Repair states---extensive debugging effort. The agent successfully built all three system configurations (baseline, rewrite-only, and integrated), implemented retrieval metrics, BERTScore, and statistical testing. But it did not complete response accuracy metrics, repeated independent runs, ablation study, query-rewriting trigger conditions, or qualitative analysis. The high Repair count suggests that the implementation was difficult (possibly due to environment issues or complex dependencies), and the repair effort consumed the budget before empirical breadth requirements could be addressed.

\textbf{idea\_14} (Adaptive FactCC Decoding, score 0.667, 6/9 items met). The Meta-Trace reveals an interesting semantic mismatch. The task text loosely described FactCC as extracting ``fact triples,'' but the actual FactCC tool is a weakly supervised document-sentence factual consistency classifier built on BERT. The agent identified this discrepancy during its Explore phase and correctly implemented FactCC-like functionality, but the scorer judged the ``FactCC Evaluation Module Implementation'' rubric item as not met, likely because the implementation diverged from the task text's literal description. The other missed items (ablation study and abstractiveness measurement) reflect the typical empirical-breadth shortfall.

\subsection{The Three Step-Limited Samples}

Three of the 40~samples reached \texttt{step\_limit\_reached} status: idea\_5 (40~steps, score 1.0), idea\_50 (60~steps, score 0.833), and idea\_10 (75~steps, score 0.667). These three form a descriptive comparison showing how step limits interact with final scores.

\textbf{idea\_5} (Dynamic Commonsense Integration, score 1.0, 40~steps) reached the step limit but achieved a perfect score. The Meta-Trace shows the system was still in its End-state verification protocol at step~40---the research work itself was already complete, and the step limit was triggered only because the multi-perspective verification protocol had not completed all rounds. The observed states are 16~Process, 11~Repair, 7~End, 2~Evaluate, 2~Review, 1~Plan, and 1~Verify. Its perfect score shows that all 12 scored items were judged met despite the status label.

\textbf{idea\_50} (Augmented Graph-Based Schema Encoding, score 0.833, 60~steps) had two unmet items (SGD-X Dataset and Repeated Runs) when the step limit was reached. The observed states are 18~Process, 15~Plan, 10~Repair, 8~Reflect, 6~End, 2~Verify, and 1~Review. The two unmet items were SGD-X dataset implementation and multiple seeds; the counts do not by themselves establish why they remained unmet.

\textbf{idea\_10} (Contextual Feedback Integration, score 0.667, 75~steps) is the case where the step limit coincided with a low score. With 75~steps, 22~Repair states, and 16~End states, this sample exhibited persistent implementation difficulties---dependency conflicts, import errors, and data format issues arose repeatedly. Unlike idea\_5, the core implementation itself was problematic, and additional steps without a strategy change may not have addressed the recurring problems; the counterfactual effect of more steps is unknown.

The descriptive comparison shows that \texttt{step\_limit\_reached} is not synonymous with a low score: idea\_5 scored 1.0, idea\_50 scored 0.833, and idea\_10 scored 0.667. Because the three tasks differ, the archive cannot isolate a causal effect of the step limit.

\subsection{The Verification--Breadth Tradeoff}

A distinctive finding from the Meta-Trace analysis is that higher repair effort is weakly \textit{positively} correlated with score ($r = 0.24$ for repair fraction vs.\ score), rather than negatively correlated as one might expect if repair consumed resources that could otherwise be used for broader experimental coverage.

\begin{table}[htbp]
\centering
\caption{Samples with the highest Repair state counts, showing the relationship between repair effort and final score. High repair does not predict low scores---\texttt{idea\_42} (25 repairs) and \texttt{idea\_5} (11 repairs) both scored 1.0.}
\label{tab:high-repair}
\small
\begin{tabularx}{\textwidth}{L{1.5cm}L{2.8cm}cccccY}
\toprule
\textbf{ID} & \textbf{Task Name} & \textbf{Score} & \textbf{Repair} & \textbf{Verify} & \textbf{End} & \textbf{Steps} & \textbf{Outcome} \\
\midrule
idea\_42 & FactCC-CaPE Integration & 1.00 & 25 & 1 & 5 & 79 & All 16 items met despite heavy repair \\
idea\_10 & Contextual Feedback & 0.67 & 22 & 0 & 16 & 75 & 5 items missed; budget consumed by repair \\
idea\_9 & Hybrid Bottleneck-Selector & 0.73 & 14 & 5 & 5 & 43 & 4 items missed; complex dependencies \\
idea\_5 & Dynamic Commonsense & 1.00 & 11 & 1 & 7 & 40 & All 12 items met; CUDA fix included \\
idea\_50 & Augmented Graph Schema & 0.83 & 10 & 2 & 6 & 60 & 2 items missed; step limit reached \\
idea\_1 & Hierarchical Hybrid QA & 0.83 & 8 & 6 & 4 & 55 & 2 items missed; high verify effort \\
idea\_19 & Cooperative Ontological NER & 0.91 & 8 & 4 & 4 & 15 & 1 item missed; efficient despite repairs \\
\bottomrule
\end{tabularx}
\end{table}

\Cref{tab:high-repair} reveals two distinct patterns among high-repair samples:

\textbf{High-repair complete runs}: \texttt{idea\_42} (25~Repair states, score 1.0) and \texttt{idea\_5} (11~Repair states, score 1.0) show that high Repair counts are compatible with complete rubric coverage. In \texttt{idea\_42}, the Meta-Trace records repairs addressing training failures, dataset format mismatches, and parameter configuration issues for the dual-dataset evaluation. The archive supports an observational reading: the run spent many steps resolving implementation and configuration issues before all 16~rubric items were judged satisfied.

\textbf{High-repair incomplete run}: \texttt{idea\_10} (22~Repair states, score 0.667) represents a contrasting outcome. Here, the repairs were concentrated on making the core system functional (handling dependency conflicts, fixing import errors, resolving data format issues), while empirical breadth requirements remained unmet in the final scored artifacts. The 16~End states (unusually high) show repeated attempts to conclude followed by further unresolved issues.

The quantitative picture is nuanced: repair fraction correlates positively with score ($r = 0.24$), but repair count does not correlate with the number of zero-scored items ($r = -0.06$). Repair states occur in both complete and incomplete runs; these observed correlations do not isolate whether repair improved scores or merely co-occurred with task and process differences.

The archive therefore supports a narrower operational lesson: Repair count alone is not a failure indicator. More targeted distinction between environmental and implementation problems may reduce unproductive cycles, but its effect requires controlled evaluation.

\subsection{Workflow State Signatures}

The aggregate Meta-Trace state distribution across the 40 runs reveals several notable patterns:

\textbf{Repair as a substantial fraction of work}: 184~Repair states constitute 18\% of all states, indicating that the system frequently detected and fixed problems during execution. This is a natural consequence of working with real code, real data, and real APIs---runtime errors, dependency conflicts, and unexpected data formats require ongoing repair.

\textbf{Repeated final checks}: 188~End states across 40~runs (mean 4.7 per run) record repeated attempts to bring a run to a verified close. In 32~of 40 runs, the last parsed Meta-Trace state was End. These are system-generated process records and should not be interpreted as independent quality judgments.

\textbf{Balanced Plan/Verify/Review/Reflect}: 79~Plan, 82~Verify, 52~Review, and 36~Reflect states indicate that the system regularly pauses to plan next steps, verify intermediate results, review accumulated evidence, and reconsider its approach. These metacognitive states complement the forward-working Process state and the reactive Repair state.

\subsection{Task Diversity}

\begin{figure}[htbp]
\centering
\includegraphics[width=0.88\textwidth]{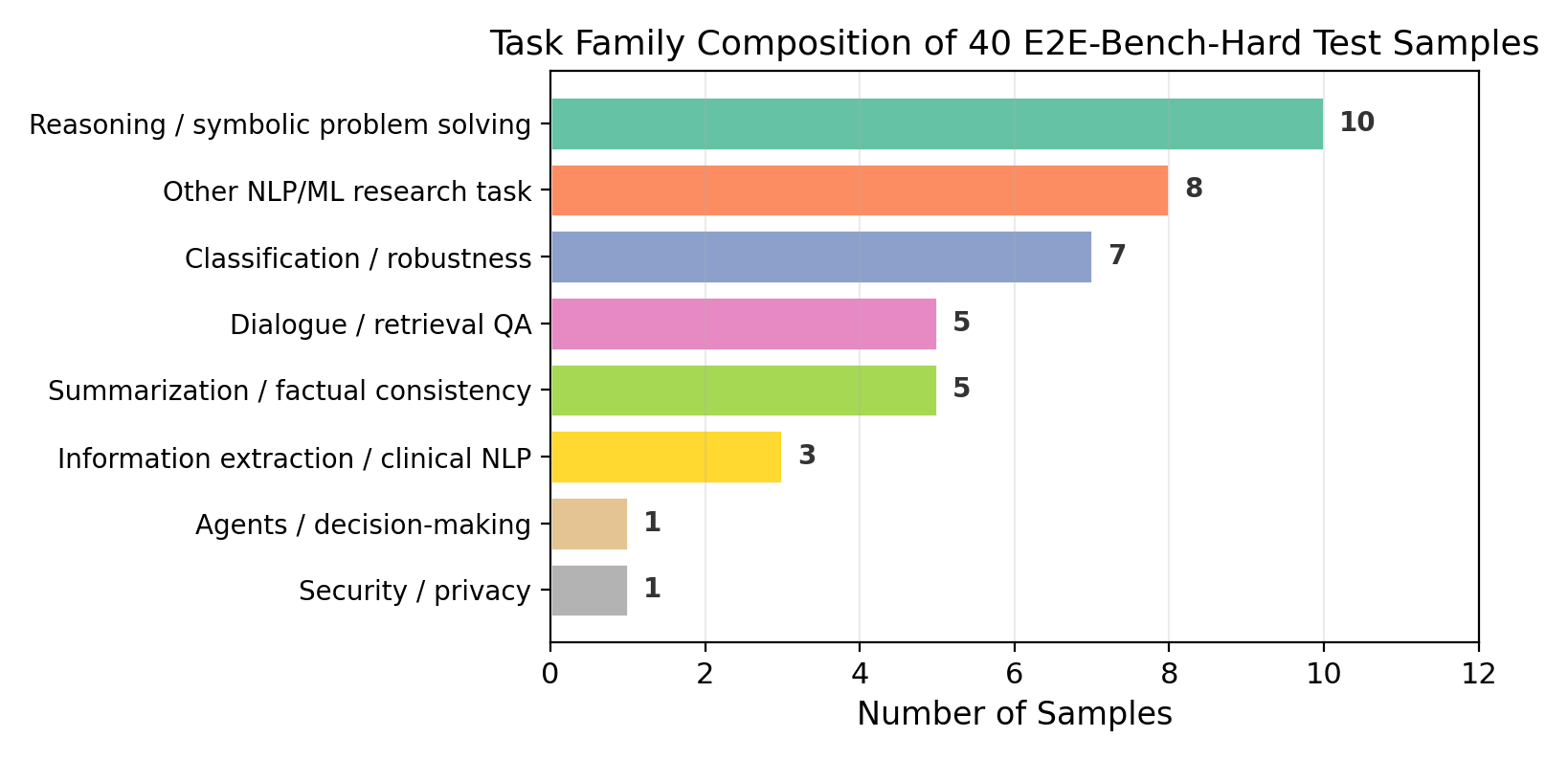}
\caption{Primary task family distribution across the 40 E2E-Bench-Hard test samples. The tasks span a broad range of AI/NLP research areas, with no single family dominating the test set.}
\label{fig:task-families}
\end{figure}

The primary-family assignment contains reasoning/symbolic problem solving (10), other NLP/ML research (8), classification/robustness (7), dialogue/retrieval QA (5), summarization/factual consistency (5), information extraction/clinical NLP (3), agents/decision-making (1), and security/privacy (1). These categories are analytical groupings of the 40 tasks rather than benchmark-defined strata.

\section{Comparison with the Official AstaBench Leaderboard}
\label{sec:comparison}

The comparison below uses the official leaderboard snapshot accessed on 4 September 2026.\cite{astabenchleaderboard}

\begin{figure}[htbp]
\centering
\includegraphics[width=\textwidth]{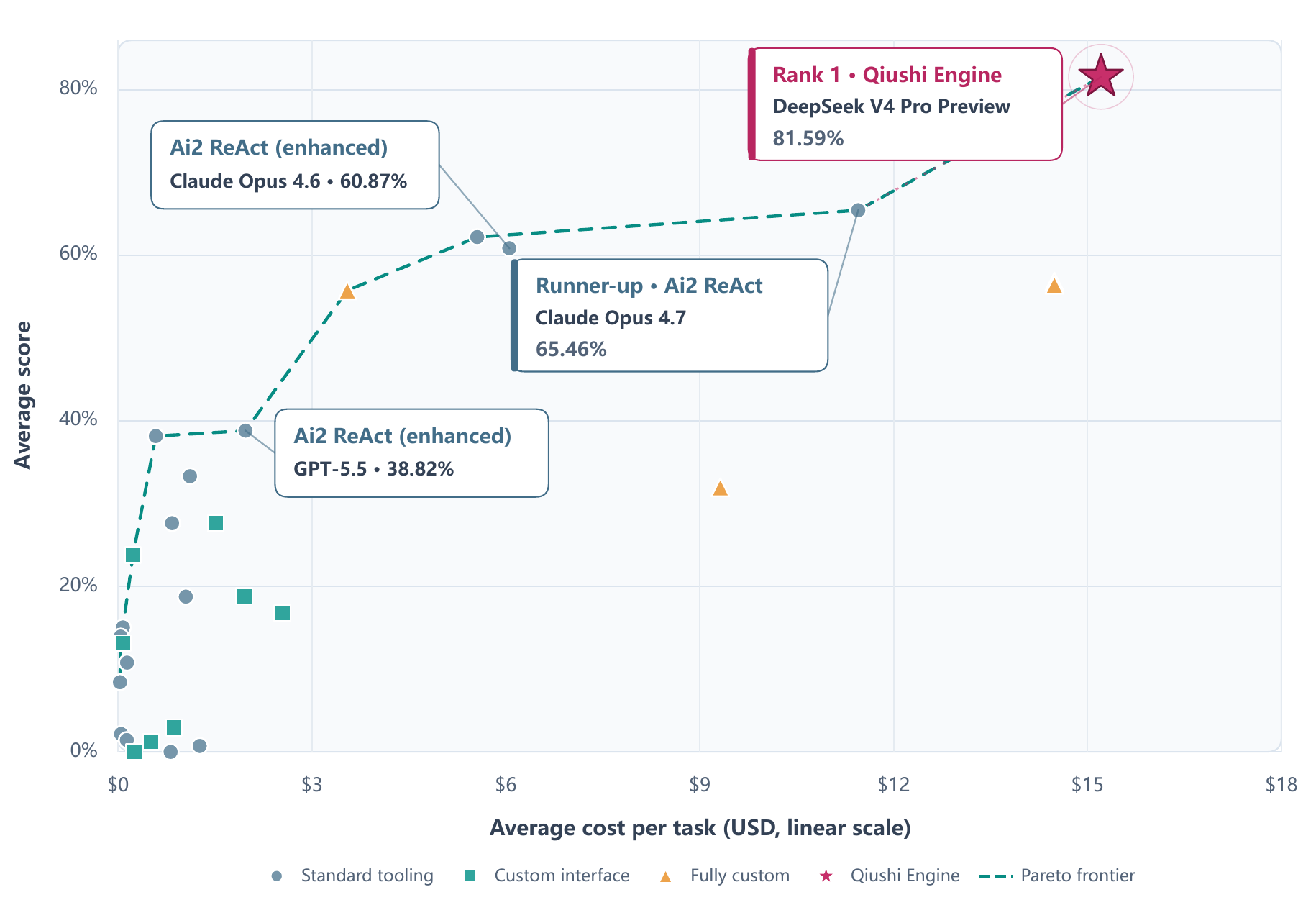}
\caption{AstaBench E2E-Bench-Hard score versus average cost per task, using the 29 results in the report's HTML snapshot of 4 September 2026. Labels retain the snapshot's two-decimal percentages; the table uses rounded leaderboard display values. All labeled Ai2 ReAct entries use Ai2's enhanced implementation; the dashed curve marks the Pareto frontier.}
\label{fig:leaderboard-linear}
\end{figure}

\Cref{fig:leaderboard-linear} shows the score--cost distribution on a linear cost axis. It highlights Qiushi Engine and Ai2's customized ReAct configurations using Claude Opus~4.7, Claude Opus~4.6, and GPT-5.5.

\begin{table}[htbp]
\centering
\caption{Official AstaBench E2E-Bench-Hard leaderboard snapshot including Qiushi Engine. Scores and costs are the values displayed by the leaderboard; score percentages are obtained by multiplying the displayed decimal by 100.}
\label{tab:comparison}
\small
\begin{tabularx}{\textwidth}{L{3.0cm}L{4.8cm}cL{2.3cm}}
\toprule
\textbf{Agent} & \textbf{Model used} & \textbf{Score} & \textbf{Cost/task} \\
\midrule
\textbf{Qiushi Engine} & \textbf{deepseek-v4pro-preview} & \textbf{81.6\%} & \textbf{\$15.209} \\
ReAct (Ai2 customized) & claude-opus-4-7 & 65.5\% & \$11.452 \\
ReAct (Ai2 customized) & claude-sonnet-4-6 & 62.2\% & \$5.557 \\
ReAct (Ai2 customized) & claude-opus-4-6 & 60.9\% & \$6.053 \\
Asta v0 & Claude Sonnet 4 (2025-05) (+6) & 56.8\% & \$14.487 \\
Asta Panda & Claude Sonnet 4 (2025-05) & 56.5\% & \$14.487 \\
Asta CodeScientist & Claude 3.7 Sonnet (2025-02) & 55.8\% & \$3.549 \\
ReAct (Ai2 customized) & GPT-5.5 (2026-04-23) & 38.8\% & \$1.971 \\
ReAct (Ai2 customized) & GPT-5 (2025-08) & 38.2\% & \$0.584 \\
\bottomrule
\end{tabularx}
\end{table}

\subsection{Score Comparison}

The official leaderboard records Qiushi Engine at 81.6\%, compared with 65.5\% for the next-highest displayed entry, Ai2's customized ReAct configuration using Claude Opus~4.7 (\Cref{tab:comparison}). The displayed difference is 16.1 percentage points. This is a leaderboard-level comparison of complete system configurations; it does not isolate the contributions of the orchestration system, model backend, tools, or execution environment.

For context, the Faker baseline---which simply prompts an LLM to fabricate research outputs without actually doing the work---scores $25.4 \pm 4.5$. This means that roughly a quarter of rubric items can be satisfied by plausible-sounding but fabricated text. The Faker comparison is an informative reference for the scoring scale, but the 56.2-point difference combines changes in model, system, tools, and execution; it is not an isolated estimate of an execution effect.

The leaderboard also shows that results vary substantially across model and agent configurations. The ReAct entries shown here are Ai2-customized implementations rather than an unmodified textbook ReAct baseline, so the table names them explicitly to avoid implying a purely generic implementation.

\subsection{Comparison Caveats}

Several caveats apply to this comparison:

\textbf{Execution environment differences.} Qiushi's submission category is ``Closed source \& UI only'' with fully custom tools, while the AstaBench agents use the standard Asta Environment with its specific code execution sandbox and tool suite. The environments therefore differ in available packages, compute resources, network access, and API availability.

\textbf{Official cost.} The reported Qiushi Engine cost is the official leaderboard's standardized average of \$15.209 per task, as used in \Cref{tab:comparison} and \Cref{fig:leaderboard-linear}.

\textbf{Scorer model.} All evaluations (both published agents and Qiushi) use the same LLM-as-judge scorer model (\texttt{claude-sonnet-4-6}), ensuring that the rubric evaluation standard is consistent.

\section{Discussion}
\label{sec:discussion}

\subsection{What the Score Means}

Let $m_i$ and $n_i$ denote met and scored items for sample $i$. The official result is the macro-average
\[
\bar s=\frac{1}{40}\sum_{i=1}^{40}\frac{m_i}{n_i}=0.815940.
\]
Pooling all items instead gives the distinct micro-average $\sum_i m_i/\sum_i n_i=416/507=0.820513$. The macro-average weights each task equally; the micro-average weights tasks with more rubric items more heavily. The rubric items are not subjective quality judgments---they are specific, binary checks on whether concrete research actions (implementing a dataset loader, running statistical tests, producing an ablation study) are supported by the combined report, code, and artifact evidence.

It is important to note what the score does \textit{not} measure. It does not assess the scientific novelty of the research findings. It does not evaluate whether the conclusions are interesting or surprising. It does not measure the quality of writing or the depth of scientific interpretation. What it measures is research-pipeline completeness: did the agent actually do the work specified in the task, and can that work be verified from the produced artifacts?

Satisfying 82\% of such items across diverse, unsimplified research tasks represents substantial coverage of the specified research pipeline. The remaining 18\% of missed items concentrate in specific categories (statistical support, external dependencies, and ablation analyses), providing a concrete view of current limitations.

\subsection{The Role of Verification and Repair}

A distinctive feature of the Qiushi process, visible in the Meta-Trace records, is the emphasis on verification and repair. The 184~Repair states and repeated End checks are aligned with the scoring demand for cross-artifact evidence.

Many rubric items require not just that code exists, but that it runs correctly and produces outputs consistent with the report. The recorded verification runs and cross-artifact checks address this requirement. In the perfect-scoring samples, the final Meta-Trace entries describe specific actions: SHA-256 hash checking of code files (\texttt{idea\_22}), determinism testing (\texttt{idea\_5}), matched-alpha control verification (\texttt{idea\_42}), and independent result reconstruction (\texttt{idea\_20}).

The verification overhead is not free---it consumes agent steps that could otherwise be used for additional experiments or analyses. This creates a resource allocation tradeoff: more verification increases the reliability of existing results but may leave less budget for expanding experimental coverage (e.g., repeated runs or ablation studies). The zero-scored dimension analysis suggests that this tradeoff is currently resolved in favor of verification over breadth, which may explain why the most commonly missed items are empirical breadth requirements rather than core implementation items.

\subsection{Score-Loss Mechanisms}

The analysis identifies four primary score-loss mechanisms:

\textbf{Resource-allocation pressure.} Statistical support, repeated runs, and ablation studies are typically the last items attempted in a research pipeline. When the core implementation and primary metrics consume most of the step budget, these breadth items are truncated. This mechanism explains the most common zero-scored category (16 samples missing statistical support).

\textbf{External dependency failures.} When a task requires specific models (GPT-3.5, LLaMA-2) or specific datasets that are unavailable in the execution environment, a cascade of downstream items---all dependent on the missing resource---score zero simultaneously. This mechanism produces the largest per-sample score drops (e.g., \texttt{idea\_16} losing 6 items from two missing models).

\textbf{Technical misinterpretation.} In rare cases, the agent's interpretation of a technical requirement differs from what the scorer expects. The \texttt{idea\_14} case (FactCC interpreted as a document-sentence classifier rather than a fact-triple extractor) illustrates this mechanism.

\textbf{Premature conclusion.} Some runs (\texttt{idea\_12} with only 13 steps) appear to conclude before the full rubric is addressed, possibly due to the agent's assessment that the task is complete when it is not. This mechanism is distinct from resource exhaustion---the budget was available but not used.

Each score-loss mechanism's relative impact can be quantified from the rubric item data. Resource-allocation pressure affected 16 of 40~samples (40\%) but typically caused only 1--2 rubric items to fail per sample, with a moderate impact on the total score (approximately 0.06--0.12 points per sample). External dependency failures affected fewer samples but more severely---idea\_16 lost 6 rubric items from two unavailable models, dropping from a potential ${\sim}1.0$ to 0.538. Technical misinterpretation affected only individual samples (idea\_14) but is difficult to prevent systematically. Premature conclusion (idea\_12, only 13 steps) affected the fewest samples but produced the largest per-sample loss. The run ended with six unmet items; the archive cannot determine whether additional steps would have improved the score.

\subsection{Relationship to Prior Research Agent Evaluations}

This evaluation's results can be understood in the broader context of AI scientific research agent assessment. The AI Scientist demonstrates the technical feasibility of automating a complete research loop~\cite{lu2024aiscientist}, while results from ScienceAgentBench, CORE-Bench, and PaperBench show that substantial gaps remain on realistic scientific coding, computational reproduction, and paper-level replication tasks~\cite{chen2024scienceagentbench,siegel2024corebench,starace2025paperbench}. AstaBench's E2E-Bench-Hard provides a more granular measurement through rubric-item scoring---rather than simply judging a paper as good or bad, it checks item by item whether specific research actions were completed.

At this granularity, Qiushi Engine's 82\% rubric-item pass rate means that across 40~diverse research tasks, the vast majority of required concrete research actions (implementing methods, constructing baselines, computing metrics, producing artifacts) were successfully executed. The concentration of failures in specific categories (statistical support, external dependencies, ablation) rather than uniform distribution indicates that the system's capability boundary is sharp rather than diffuse.

The Faker baseline (25.4) provides an important reference: approximately one quarter of rubric items can be satisfied by plausible-sounding but fabricated text. The 56.2~percentage-point gap between Qiushi (81.6) and Faker (25.4) marks the scale difference between fabricated outputs and a run that produced code, data processing, experiment execution, and artifacts, while still combining model, system, tool, and execution differences.

\subsection{System-Level Insights}

This evaluation provides several workflow-design insights:

\textbf{Value of separating research functions.} Keeping planning, method construction, experiment execution, and final checking as explicit functions is consistent with E2E-Bench-Hard's demand for both working code and cross-artifact evidence. The observed archive supports the presence of this separation, but the evaluation does not isolate its causal contribution to the score.

\textbf{Repair as a research capability.} The 184~Repair states should not be viewed as a system deficiency. In real research, runtime errors, dependency conflicts, data format mismatches, and unexpected behavior are normal engineering realities. The system's ability to detect these problems and fix them automatically---rather than ignoring or hiding them---is an observable feature of these runs. The weak positive correlation between repair fraction and score ($r = 0.24$) is descriptive; it does not establish that repair caused higher scores.

\textbf{Meta-Trace as process evidence.} The Meta-Trace memory system serves as runtime long-horizon memory and provides the process evidence analyzed in this report. Its reasoning summaries and structured handoffs create an inspectable process--outcome record. This inspectability has independent value in AI system evaluation: it allows evaluators to understand not just what the score was, but how it was produced.

\textbf{DeepSeek v4 Pro Preview in this configuration.} All 40~runs used DeepSeek \texttt{deepseek-v4pro-preview}. The official 81.6\% result applies to the complete Qiushi--model--tool configuration and should not be attributed to the model alone. Qiushi Engine's backend is configurable; DeepSeek was the choice for this evaluation, not a product constraint. Controlled backend comparisons remain future work.

\subsection{Evidence Synthesis: From Paired Observations to Archive-Level Patterns}

The case studies, zero-scored dimension analysis, and process signature data form a multi-layer evidence structure that allows the same results to be examined from different angles.

\textbf{Case observations and global dimension patterns.} The loss patterns observed in individual cases are consistent with the archive-wide zero-scored dimension distribution. The idea\_5 vs idea\_10 comparison shows that \texttt{step\_limit\_reached} does not predict score---the global correlation between step count and score ($r = -0.01$) confirms this sample-level observation across all 40~tasks. The idea\_16 vs idea\_12 comparison distinguishes ``external dependency cascade'' from ``breadth coverage shortfall'' as distinct failure structures---the global dimension analysis similarly shows ``core method component'' (11~samples, often involving external dependencies) and ``statistical support'' (16~samples, involving breadth shortfall) as two independent principal sources of score loss. The idea\_37 case---65~steps, 84.9M tokens, 54.1~hours, yet three specific items unmet---validates the global finding that token volume does not substitute for specific rubric-item fulfillment ($r = 0.001$).

\textbf{Process signatures and scoring outcomes.} Process-state data provides descriptive evidence about workflow allocation but not causal evidence about scores. Repair states (184~total) appear in both perfect-score and constrained samples, showing that repair is a normal component of the observed research process rather than a score predictor by itself. End states (188~total, mean 4.7/run) record repeated final checks; the mean End count for perfect-score samples (5.3) is slightly higher than in other bands, consistent with but not proving that verification contributed to high coverage.

\textbf{From sample level to archive level.} The macro-average of $0.8159$ is an archive-level statistic whose standard error ($0.0187$) reflects how the substantial single-sample variance ($\sigma = 0.119$) is smoothed by 40-sample aggregation. Four perfect-score samples and four constrained samples occupy the distribution tails, but most samples (24/40 $\geq 0.80$) fall in the ``core implementation complete, marginal coverage incomplete'' range. This left-skewed, high-concentration distribution shape indicates stable system capability on core research pipelines, with marginal variation driven primarily by task-specific conditions (external dependency availability, rubric-item count and type, task complexity) rather than systematic capability deficits.

\textbf{Evidence chain completeness.} All quantitative statements in this report trace to two classes of raw data in the submission archive: scorer JSON files (containing per-item binary judgments and natural-language explanations) and Meta-Trace files (containing per-step process records). The scoring mechanism in \S\ref{sec:benchmark} and the evidence-source mapping in \S\ref{sec:submission} show the path from task specification to macro-average. All values used in paired comparisons come from \texttt{e2e\_hard\_per\_sample\_rubric\_process\_summary.csv}, which was extracted from the raw archive by automated scripts and verified through six consistency checks.

\subsection{Methodology Considerations}

Several aspects of the evaluation methodology merit explicit acknowledgment. The rubric-item scores are produced by an LLM judge (\texttt{claude-sonnet-4-6}), not by human experts; binary scoring does not capture partial completion; the official score is a macro-average over sample-level means, which differs from the micro-average item fulfillment rate (416/507 = 82.1\% vs.\ 81.6\%). The zero-scored dimension categories (statistical support, core components, etc.) are analytical classifications derived from rubric-item keywords and case inspection, not labels produced by the scorer itself; categories can overlap within a single sample. Meta-Trace is a system-generated process record, not an independent observation. UI mode, task family, and resource allocation were not randomly assigned. Execution environments, models, and tools differ across leaderboard entries, so the benchmark comparison reflects complete system configurations rather than isolating any single factor. The Qiushi score and standardized cost are now recorded on the official leaderboard.

\subsection{Opportunities for Improvement}

The zero-scored dimension analysis points to concrete improvement opportunities:

\begin{enumerate}
\item \textbf{Early statistical planning.} Prioritize statistical testing and repeated runs as integral parts of the experimental design, scheduled before the step budget is consumed by implementation and repair. Rather than treating repeated runs as an optional extension, plan for them from the first Explore step.

\item \textbf{Dependency pre-checking.} Before committing to an experimental design that requires specific external models, verify their availability in the execution environment. If a required model is unavailable, adapt the experimental design (e.g., using an available alternative with appropriate justification) rather than proceeding with a plan that cannot be completed.

\item \textbf{Rubric-aware scheduling.} Since each task comes with an explicit rubric, the Orchestrator could use the rubric items as a checklist during the Execute phase to ensure that all required dimensions are addressed before the session moves to Express.

\item \textbf{Budget reservation.} Reserve a fraction of the step budget for ablation studies and sensitivity analyses, scheduling them before final verification rather than after core implementation.
\end{enumerate}

\section{Conclusions}
\label{sec:conclusions}

Qiushi Engine v0.8, using DeepSeek \texttt{deepseek-v4pro-preview} as the selected model backend, is recorded on the official AstaBench E2E-Bench-Hard leaderboard with a score of 0.816 and a standardized cost of \$15.209 per task. The next-highest displayed entry scores 0.655, a leaderboard difference of 16.1 percentage points. In addition, Qiushi perfectly completed 4 of 40 tasks (10\%), compared with the approximately 3\% best full-task completion rate reported for AstaBench's official agents---a gain of 7 percentage points and roughly a 3.3-fold rate. These comparisons apply to complete system configurations rather than any single component.

The process analysis shows strong coverage of core research implementation---dataset handling, method and baseline construction, primary metric computation, and cross-artifact evidence bundling. Score losses concentrate in specific rubric dimensions: statistical support (16 samples), core method components requiring external resources (11), explicit metric completion (10), and ablation studies (8). These observed gaps suggest concrete improvements in resource allocation and dependency management.

AstaBench E2E-Bench-Hard tests a demanding form of automated research capability: completing unsimplified AI/NLP research cycles from task specification to evidence-supported technical reports. Qiushi's result provides evidence of broad pipeline coverage under this benchmark, while the rubric analysis also makes clear that scientific novelty, human expert judgment, and causal attribution of system components require separate evaluation.

\appendix
\begin{landscape}
\section{Full 40-Sample Evidence Table}
\label{app:samples}

\Cref{tab:all-samples} lists all 40 test samples sorted by score, with scored item counts, zero-scored dimensions, task family, process length, resource usage, and completion status.

\rowcolors{2}{ReportLight}{white}
{\scriptsize
\setlength{\tabcolsep}{2.5pt}
\renewcommand{\arraystretch}{1.12}
\begin{longtable}{L{1.25cm}L{4.6cm}cL{1.1cm}cccL{7.1cm}L{5.0cm}}
\caption{All 40 E2E-Bench-Hard test samples sorted by score. The unmet-items column lists up to two representative items; gap categories use readable short labels.}\label{tab:all-samples}\\
\toprule
\textbf{ID} & \textbf{Task name} & \textbf{Score} & \textbf{Met/Tot} & \textbf{Steps} & \textbf{Tok(M)} & \textbf{Hours} & \textbf{Representative unmet items} & \textbf{Gap categories} \\
\midrule
\endfirsthead
\multicolumn{9}{l}{\small\textit{(continued from previous page)}}\\
\toprule
\textbf{ID} & \textbf{Task name} & \textbf{Score} & \textbf{Met/Tot} & \textbf{Steps} & \textbf{Tok(M)} & \textbf{Hours} & \textbf{Representative unmet items} & \textbf{Gap categories} \\
\midrule
\endhead
\bottomrule
\endlastfoot
idea\_12 & TAPP with Explicit Demonstrations & 0.50 & 6/12 & 13 & 10.2 & 4.5 & Large Language Model Selection; Accuracy Measurement; ... & metrics, statistics, ablation/sensitivity, required model \\
idea\_16 & Anticipatory Ensemble for Medication Extraction & 0.54 & 7/13 & 38 & 52.8 & 13.5 & GPT-3.5 Model Integration; LLaMA-2 Model Integration; ... & statistics, required model, core method \\
idea\_10 & Contextual Feedback Integration & 0.67 & 10/15 & 75 & 103.2 & 28.9 & Response Accuracy Metrics; Multiple Experimental Runs; ... & metrics, statistics, ablation/sensitivity, qualitative evaluation \\
idea\_14 & Adaptive FactCC Decoding & 0.67 & 6/9 & 15 & 13.2 & 1.8 & FactCC Evaluation Module Implementation; Ablation Study; ... & metrics, ablation/sensitivity, core method \\
idea\_38 & Tree-Logical Integration & 0.70 & 7/10 & 15 & 9.3 & 0.9 & Statistical Comparison; Multiple Experimental Runs; ... & statistics, required model \\
idea\_39 & RCI-Peer Review Integration & 0.70 & 7/10 & 28 & 27.8 & 16.7 & Secondary Metric: Reasoning Steps; Multiple Independent Runs; ... & metrics, statistics, qualitative evaluation \\
idea\_31 & Integrated Rule-Entailment Prototypical Networks & 0.71 & 10/14 & 21 & 32.6 & 12.6 & TACRED Dataset Implementation; Attention Mechanism Integration; ... & ablation/sensitivity, generalization/scaling, core method \\
idea\_15 & mT5 Cross-Lingual Integration & 0.73 & 8/11 & 15 & 19.3 & 4.5 & X-Fact Dataset Loading; Low-Resource Language Analysis; ... & statistics \\
idea\_37 & Reframed Coherent Instructions & 0.73 & 8/11 & 65 & 84.9 & 54.1 & F1 Score Calculation; Reframing Technique Documentation; ... & other \\
idea\_47 & Universal RandAugment Transformers & 0.73 & 8/11 & 10 & 8.2 & 1.2 & Compositional Generalization Evaluation; Multiple Training Runs; ... & ablation/sensitivity, generalization/scaling \\
idea\_9 & Hybrid Bottleneck-Selector Integration & 0.73 & 11/15 & 43 & 62.3 & 8.1 & Baseline System 2: Neural Selector Only; Query Throughput Measurement; ... & metrics, statistics, core method \\
idea\_24 & Entity-Prune Summarization & 0.75 & 9/12 & 29 & 34.5 & 8.5 & Dataset Selection and Preparation; Hallucination Rate Evaluation; ... & other \\
idea\_27 & Adversarial Signal Transformation & 0.75 & 9/12 & 14 & 11.7 & 1.8 & Speaker Diversity Testing; Ablation Study; ... & ablation/sensitivity \\
idea\_34 & AlignScore-NLI Integration & 0.75 & 6/8 & 12 & 9.6 & 1.2 & Text Generation Model; AlignScore Implementation & core method \\
idea\_23 & Dynamic Task Weighting in Multi-Task Fact Verification & 0.77 & 10/13 & 27 & 36.6 & 8.3 & Health Domain Dataset; Statistical Significance Testing; ... & statistics \\
idea\_35 & Constraint-CoT Integration & 0.79 & 11/14 & 19 & 18.6 & 1.2 & Task Completion Time Measurement; Success Rate Measurement; ... & metrics \\
idea\_7 & Cross-Document Structured Extraction & 0.80 & 12/15 & 45 & 66.3 & 5.9 & Dataset Acquisition; Statistical Significance Testing; ... & statistics \\
idea\_30 & Cluster-Enhanced Dimensional Reduction & 0.80 & 12/15 & 15 & 17.5 & 11.6 & Hyperparameter Optimization; Isotropy Measurement; ... & metrics, ablation/sensitivity \\
idea\_43 & Integrated Verification and Critic Framework & 0.80 & 8/10 & 26 & 32.8 & 5.4 & Knowledge Graph Integration; Automated Experimentation Framework & core method \\
idea\_1 & Hierarchical Hybrid QA Integration & 0.83 & 10/12 & 55 & 81.1 & 10.9 & Hybrid Evidence Utilization Algorithm; Multiple Experimental Runs & statistics \\
idea\_11 & Offline-Continuous Prompt Optimization & 0.83 & 10/12 & 27 & 35.3 & 7.2 & Secondary Metrics; Containerized Environment & metrics \\
idea\_46 & CoT-Enhanced Type Annotations & 0.83 & 10/12 & 18 & 9.4 & 1.6 & Type Correctness Evaluation; Multiple Experimental Runs & statistics \\
idea\_50 & Augmented Graph-Based Schema Encoding & 0.83 & 10/12 & 60 & 79.9 & 12.7 & SGD-X Dataset Implementation; Multiple Runs with Different Seeds & core method \\
idea\_13 & DEBIE-MOMA Integration for Bias Mitigation & 0.85 & 11/13 & 14 & 10.8 & 1.3 & DEBIE Platform Implementation; MOMA Framework Implementation & core method \\
idea\_4 & Emotion-Enhanced Multimodal CoT & 0.86 & 12/14 & 28 & 49.9 & 11.8 & Vision-Language Transformer; Emotion Classification Accuracy Evaluation & other \\
idea\_48 & Mix Self-Consistency with Self-Refinement & 0.86 & 12/14 & 24 & 19.1 & 4.0 & Consistency Rate Measurement; Multiple Experimental Runs & metrics, statistics \\
idea\_25 & Integrated Model Editing for Consistency & 0.87 & 13/15 & 39 & 53.3 & 49.9 & Multiple Independent Runs; Meta-Analysis & statistics \\
idea\_17 & Diversified Causal Reasoning & 0.88 & 14/16 & 16 & 13.3 & 6.0 & Multiple Runs for Statistical Confidence; GPT-4 Model Implementation & statistics, core method \\
idea\_19 & Cooperative Ontological NER & 0.91 & 10/11 & 15 & 12.7 & 1.4 & Standard NER Baseline & core method \\
idea\_21 & Integrated Feedback and Self-Correction & 0.91 & 10/11 & 27 & 37.9 & 8.3 & Multiple Experimental Runs & statistics \\
idea\_32 & Dynamic Streaming Evaluation & 0.92 & 12/13 & 10 & 8.6 & 1.4 & Computational Efficiency Measurement & metrics, efficiency/cost \\
idea\_33 & Adversarial Contextual Embeddings & 0.92 & 12/13 & 13 & 11.9 & 24.3 & Coded Hate Speech Detection Analysis & other \\
idea\_26 & Integrated Contrastive Learning for Factual Extraction & 0.93 & 13/14 & 20 & 28.0 & 11.6 & Multiple Experimental Runs & statistics \\
idea\_41 & FastText-Enhanced Membership Inference & 0.93 & 13/14 & 25 & 38.2 & 5.4 & Multiple Dataset Validation & other \\
idea\_28 & Contrastive Bayesian Hybrid Learning & 0.93 & 14/15 & 14 & 10.5 & 1.1 & Low-Resource Dataset Implementation & core method \\
idea\_20 & Syntactic Trigger Inspection & 0.94 & 16/17 & 17 & 24.0 & 3.0 & Varying Poisoning Rates & ablation/sensitivity \\
idea\_5 & Dynamic Commonsense Integration & 1.00 & 12/12 & 40 & 56.5 & 6.2 & --- & none \\
idea\_22 & Memory-Augmented Decentralized Decision-Making & 1.00 & 10/10 & 20 & 15.2 & 2.5 & --- & none \\
idea\_29 & Memory-Enhanced Tree Reasoning & 1.00 & 11/11 & 20 & 24.1 & 3.7 & --- & none \\
idea\_42 & FactCC-CaPE Integration & 1.00 & 16/16 & 79 & 105.4 & 21.1 & --- & none \\
\end{longtable}
}
\end{landscape}

\section{Rubric Dimension Inventory}
\label{app:rubric-inventory}

Across the 40~test samples, the 507~scored rubric items were classified into dimension categories based on the research action they require. \Cref{tab:rubric-inventory} shows the categories, the number of samples affected by zero-scored items in each category, the primary mechanism that causes failures, and representative examples from the archive.

\begin{table}[htbp]
\centering
\caption{Rubric dimension categories with zero-scored sample counts, primary failure mechanisms, and representative examples.}
\label{tab:rubric-inventory}
\small
\begin{tabularx}{\textwidth}{L{3.2cm}cL{3cm}L{4.5cm}}
\toprule
\textbf{Dimension Category} & \textbf{Samples} & \textbf{Primary Cause} & \textbf{Representative Examples} \\
\midrule
Statistical support / repeated runs & 16 & Resource-allocation pressure & ``Multiple Experimental Runs'' in idea\_35; ``Statistical Significance Testing'' in idea\_7, idea\_30, idea\_23 \\
Core method component & 11 & External dependency or implementation complexity & ``Neural Selector'' in idea\_9; ``KG Integration'' in idea\_43; ``FactCC Eval Module'' in idea\_14 \\
Required metrics not completed & 10 & Implementation gaps or metric misinterpretation & ``Response Accuracy'' in idea\_10; ``Type Correctness'' in idea\_46; ``Consistency Rate'' in idea\_48 \\
Ablation / sensitivity & 8 & Resource-allocation pressure; late scheduling & ``Ablation Study'' in idea\_10, idea\_14, idea\_27; ``Hyperparameter'' in idea\_27, idea\_47 \\
Other specific requirements & 5 & Task-specific constraints & ``Compositional Generalization'' in idea\_47; ``Hallucination Rate'' in idea\_24 \\
Exact model / backbone & 3 & External dependency failure & ``GPT-3.5 Integration'' in idea\_16; ``LLaMA-2 Integration'' in idea\_16; ``LLM Base'' in idea\_38 \\
Human / qualitative eval & 2 & Inherently difficult for automated agents & ``Qualitative Analysis'' in idea\_10; ``Peer-Review Criteria'' in idea\_39 \\
Generalization / scaling & 2 & Resource-allocation pressure & ``Cross-Lingual Evaluation'' in idea\_34; ``Compositional Gen.'' in idea\_47 \\
Efficiency analysis & 1 & Low priority under budget constraints & (none in constrained set; 1~sample had efficiency-related optional item) \\
\bottomrule
\end{tabularx}
\end{table}

The distribution reveals a clear hierarchy of failure mechanisms. Resource-allocation pressure accounts for the largest share of failures (statistical support, ablation, generalization, efficiency), affecting items that are typically scheduled last in a research pipeline. External dependency failures produce the most severe per-sample impact (idea\_16 losing 6~items from two unavailable models), but affect fewer samples overall. Implementation complexity and metric misinterpretation fall between these extremes.

A notable feature is that no sample scored zero on \textit{all} implementation items---even the lowest-scoring sample (\texttt{idea\_12} at 0.50) satisfied 6 of 12 items, including the TAPP framework implementation, both datasets, and the prompt format documentation. This is consistent with high observed coverage of implementation-related items in this archive; the zero-scored items cluster around empirical breadth rather than core execution.

\section{Submission Archive Contents}
\label{app:archive}

The submission archive (\texttt{qiushi\_engine\_e2e\_bench\_hard\_test\_ui\_only\_v0.8\_20260824}) contains the following files:

\begin{itemize}
\item \texttt{qiushi\_engine\_e2e\_bench\_hard\_test.eval} --- Zipped Inspect log (112~MB) containing 45 entries: 40 sample JSONs (\texttt{samples/idea\_*\_epoch\_1.json}), journal start record, header, summaries, and reductions.
\item \texttt{scores.json} --- Per-sample and aggregate scores from the official \texttt{astabench score} command.
\item \texttt{summary\_stats.json} --- Task-level statistics including the E2E-Bench-Hard test score (0.8159401260504202) and standard error (0.018745023003278058).
\item \texttt{submission.json} --- Submission metadata: agent name (Qiushi Engine), version (v0.8), actual model (deepseek-v4pro-preview), openness (Closed source \& UI only), tool usage (Fully custom).
\item \qpath{test_model_configuration.json} --- Model-configuration record (\Cref{tab:model-identity}).
\item \texttt{test\_ui\_telemetry.json} --- Per-sample telemetry: session IDs, steps, turns, tools, LLM calls, duration, and token counts.
\item \texttt{source\_log\_inventory.json} --- Per-sample score, run label, and SHA-256 hashes for packaged results and source eval logs.
\item \texttt{eval\_config.json} --- Inspect evaluation configuration.
\item \texttt{SHA256SUMS.json}, \texttt{SHA256SUMS.txt} --- Integrity checksums.
\item \texttt{README.md} --- Archive documentation with replay methodology.
\end{itemize}

\begin{landscape}
\section{Per-Sample Process Signature Details}
\label{app:process-details}

\Cref{tab:process-details} provides the per-sample Meta-Trace state counts for all 40~samples, enabling detailed process analysis. The states are: Process (P, forward work), Repair (R, fixing problems), End (E, final verification), Verify (V, checking intermediate results), Plan (Pl, making or repairing plans), Review (Rv, checking quality), Reflect (Rf, reconsidering approach), and Evaluate (Ev, comparing choices).

\rowcolors{2}{ReportLight}{white}
{\scriptsize
\setlength{\tabcolsep}{3pt}
\renewcommand{\arraystretch}{1.12}
\begin{longtable}{L{1.3cm}L{7.8cm}c*{8}{c}}
\caption{Per-sample Meta-Trace state counts for all 40 samples, sorted by score.}\label{tab:process-details}\\
\toprule
\textbf{ID} & \textbf{Task name} & \textbf{Score} & \textbf{P} & \textbf{R} & \textbf{E} & \textbf{V} & \textbf{Pl} & \textbf{Rv} & \textbf{Rf} & \textbf{Ev} \\
\midrule
\endfirsthead
\multicolumn{11}{l}{\small\textit{(continued from previous page)}}\\
\toprule
\textbf{ID} & \textbf{Task name} & \textbf{Score} & \textbf{P} & \textbf{R} & \textbf{E} & \textbf{V} & \textbf{Pl} & \textbf{Rv} & \textbf{Rf} & \textbf{Ev} \\
\midrule
\endhead
\bottomrule
\endlastfoot
idea\_12 & TAPP with Explicit Demonstrations & 0.50 & 1 & 1 & 6 & 1 & 1 & 1 & 2 & 0 \\
idea\_16 & Anticipatory Ensemble for Medication Extraction & 0.54 & 6 & 3 & 9 & 11 & 9 & 0 & 0 & 0 \\
idea\_10 & Contextual Feedback Integration & 0.67 & 24 & 22 & 16 & 0 & 3 & 8 & 2 & 0 \\
idea\_14 & Adaptive FactCC Decoding & 0.67 & 0 & 0 & 0 & 0 & 0 & 0 & 0 & 0 \\
idea\_38 & Tree-Logical Integration & 0.70 & 7 & 0 & 4 & 4 & 1 & 1 & 2 & 0 \\
idea\_39 & RCI-Peer Review Integration & 0.70 & 6 & 5 & 5 & 8 & 1 & 1 & 0 & 2 \\
idea\_31 & Integrated Rule-Entailment Prototypical Networks & 0.71 & 6 & 3 & 5 & 4 & 3 & 0 & 0 & 0 \\
idea\_15 & mT5 Cross-Lingual Integration & 0.73 & 4 & 3 & 4 & 0 & 2 & 1 & 1 & 0 \\
idea\_37 & Reframed Coherent Instructions & 0.73 & 48 & 5 & 4 & 0 & 3 & 1 & 0 & 4 \\
idea\_47 & Universal RandAugment Transformers & 0.73 & 0 & 0 & 0 & 0 & 0 & 0 & 0 & 0 \\
idea\_9 & Hybrid Bottleneck-Selector Integration & 0.73 & 11 & 14 & 5 & 5 & 2 & 0 & 0 & 6 \\
idea\_24 & Entity-Prune Summarization & 0.75 & 12 & 8 & 6 & 1 & 2 & 0 & 0 & 0 \\
idea\_27 & Adversarial Signal Transformation & 0.75 & 0 & 0 & 0 & 0 & 0 & 0 & 0 & 0 \\
idea\_34 & AlignScore-NLI Integration & 0.75 & 0 & 0 & 0 & 0 & 0 & 0 & 0 & 0 \\
idea\_23 & Dynamic Task Weighting in Multi-Task Fact Verification & 0.77 & 8 & 6 & 5 & 3 & 2 & 2 & 0 & 1 \\
idea\_35 & Constraint-CoT Integration & 0.79 & 6 & 3 & 5 & 3 & 0 & 1 & 1 & 0 \\
idea\_7 & Cross-Document Structured Extraction & 0.80 & 20 & 3 & 5 & 7 & 5 & 1 & 4 & 0 \\
idea\_30 & Cluster-Enhanced Dimensional Reduction & 0.80 & 4 & 0 & 9 & 0 & 1 & 1 & 0 & 0 \\
idea\_43 & Integrated Verification and Critic Framework & 0.80 & 16 & 0 & 5 & 0 & 2 & 2 & 0 & 1 \\
idea\_1 & Hierarchical Hybrid QA Integration & 0.83 & 26 & 8 & 4 & 6 & 8 & 0 & 0 & 3 \\
idea\_11 & Offline-Continuous Prompt Optimization & 0.83 & 11 & 9 & 4 & 0 & 1 & 1 & 0 & 1 \\
idea\_46 & CoT-Enhanced Type Annotations & 0.83 & 0 & 0 & 0 & 0 & 0 & 0 & 0 & 0 \\
idea\_50 & Augmented Graph-Based Schema Encoding & 0.83 & 18 & 10 & 6 & 2 & 15 & 1 & 8 & 0 \\
idea\_13 & DEBIE-MOMA Integration for Bias Mitigation & 0.85 & 0 & 0 & 0 & 0 & 0 & 0 & 0 & 0 \\
idea\_4 & Emotion-Enhanced Multimodal CoT & 0.86 & 12 & 1 & 6 & 3 & 1 & 5 & 0 & 0 \\
idea\_48 & Mix Self-Consistency with Self-Refinement & 0.86 & 5 & 6 & 5 & 2 & 1 & 4 & 1 & 0 \\
idea\_25 & Integrated Model Editing for Consistency & 0.87 & 21 & 0 & 8 & 5 & 1 & 1 & 2 & 0 \\
idea\_17 & Diversified Causal Reasoning & 0.88 & 2 & 3 & 5 & 1 & 1 & 2 & 2 & 0 \\
idea\_19 & Cooperative Ontological NER & 0.91 & 4 & 8 & 4 & 4 & 1 & 3 & 3 & 2 \\
idea\_21 & Integrated Feedback and Self-Correction & 0.91 & 3 & 7 & 9 & 1 & 7 & 0 & 0 & 0 \\
idea\_32 & Dynamic Streaming Evaluation & 0.92 & 0 & 0 & 0 & 0 & 0 & 0 & 0 & 0 \\
idea\_33 & Adversarial Contextual Embeddings & 0.92 & 0 & 1 & 5 & 1 & 2 & 2 & 1 & 0 \\
idea\_26 & Integrated Contrastive Learning for Factual Extraction & 0.93 & 12 & 0 & 5 & 0 & 1 & 2 & 0 & 0 \\
idea\_41 & FastText-Enhanced Membership Inference & 0.93 & 7 & 8 & 4 & 0 & 0 & 3 & 3 & 0 \\
idea\_28 & Contrastive Bayesian Hybrid Learning & 0.93 & 0 & 0 & 0 & 0 & 0 & 0 & 0 & 0 \\
idea\_20 & Syntactic Trigger Inspection & 0.94 & 7 & 0 & 5 & 1 & 1 & 3 & 0 & 0 \\
idea\_5 & Dynamic Commonsense Integration & 1.00 & 16 & 11 & 7 & 1 & 1 & 2 & 0 & 2 \\
idea\_22 & Memory-Augmented Decentralized Decision-Making & 1.00 & 7 & 7 & 8 & 3 & 0 & 1 & 4 & 2 \\
idea\_29 & Memory-Enhanced Tree Reasoning & 1.00 & 4 & 4 & 5 & 4 & 0 & 0 & 0 & 3 \\
idea\_42 & FactCC-CaPE Integration & 1.00 & 41 & 25 & 5 & 1 & 1 & 2 & 0 & 4 \\
\end{longtable}
}
\end{landscape}

The process signature table should be read as parsed process-state evidence. Some high-scoring runs were short (for example, \texttt{idea\_32}: 10~steps and \texttt{idea\_33}: 13~steps), but rows with all-zero state counts indicate unavailable or unparsed state evidence rather than a zero-state workflow. High-repair runs such as \texttt{idea\_42} (25~Repair, score 1.0) and \texttt{idea\_10} (22~Repair, score 0.667) show opposite scored outcomes; the process counts describe workflow allocation and do not by themselves identify causal effects on score.

\section{Data Processing Methodology}
\label{app:methodology}

Figures and tables describing individual runs were generated from the submission archive and the 40 Meta-Trace process records using reproducible Python scripts. The leaderboard comparison and reported monetary cost use the official leaderboard snapshot. The processing pipeline consisted of the following steps:

\subsection{Script Descriptions and Verification Chain}

The data processing pipeline consists of five Python scripts, each reading from the previous step's output and producing inputs for the next, forming a complete reproducible evidence chain.

\textbf{Script~1: \qpath{aggregate_e2e_hard_submission.py}.} Input: \qpath{scores.json}, \qpath{summary_stats.json}, \qpath{submission.json}, \qpath{test_model_configuration.json}, \qpath{test_ui_telemetry.json}, and \qpath{source_log_inventory.json}. Processing: merges six JSON files into per-sample records and computes aggregate statistics (mean, standard error, median, min/max, total tokens, total duration). Output: \qpath{e2e_hard_40_run_summary.csv} (40~rows, 16~columns per row) and \qpath{e2e_hard_aggregate_stats.json}. Verification: confirms sample count~=~40, score mean exactly matches the official value from \qpath{summary_stats.json}.

\textbf{Script~2: \qpath{inspect_eval_structure.py}.} Input: \qpath{qiushi_engine_e2e_bench_hard_test.eval} (112~MB zipped Inspect log). Processing: reads all 45~entries via zipfile module (1~journal, 40~samples, summaries, reductions, header), extracts header metadata (task name, model identifier, scorer name), and records each sample's event-chain structure. Output: \qpath{eval_structure_summary.json}. Verification: confirms the header task \qpath{astabench/e2e_discovery_hard_test}, model \qpath{deepseek/deepseek-v4pro-preview}, and scorer \qpath{score_rubric}.

\textbf{Script~3: \texttt{extract\_e2e\_hard\_evidence\_chain.py}.} Input: the 40~sample JSONs from the \texttt{.eval} file. Processing: extracts task definitions (name, description, hypothesis, variables, contrasts, measurements), score values, scorer explanation text, event counts, and output package sizes from each sample's Inspect events; merges with UI telemetry and Meta-Trace process summaries. Output: \texttt{e2e\_hard\_per\_sample\_evidence\_chain.csv} (10.9~MB, 40~rows with complete scorer explanation text and event chains). Verification: extracted scores compared per-sample against \texttt{source\_log\_inventory.json}, confirming zero mismatches.

\textbf{Script~4: \qpath{extract_all_rubric_process_patterns.py}.} Input: Script~3 output. Processing: parses scorer explanation strings to extract individual rubric items and corresponding binary scores; simultaneously extracts process-state counts (Process, Repair, End, Verify, Plan, Review, Reflect, Evaluate) from Meta-Trace files. Output: \qpath{e2e_hard_rubric_item_scores_long.csv} (507~rows in long format) and \qpath{e2e_hard_per_sample_rubric_process_summary.csv} (40~rows). Verification: the recomputed score mean matches the official value to all significant digits ($0.8159401260504202$); total scored items~=~507, met items~=~416, and zero-scored items~=~91.

\textbf{Script~5: \texttt{build\_report\_figures\_tables.py}.} Input: all outputs from Scripts~1--4. Processing: generates 11~analytical figures and 5~structured JSON tables. Verification: all numerical values consistent with source data; image files non-empty.

\subsection{Cross-Step Consistency Verification}

To ensure numerical accuracy throughout this report, the data processing chain performs consistency verification at two independent levels:

\textbf{Level~1: Source data internal consistency.} Per-sample scores from \texttt{scores.json} are compared one-by-one against the independent records in \texttt{source\_log\_inventory.json}, confirming exact matches for all 40~samples.

\textbf{Level~2: Extraction--recomputation consistency.} Per-rubric-item scores parsed from scorer explanation text are reaggregated into sample-level scores and compared against official scores. The recomputed mean ($0.8159401260504202$) matches the official value to all significant digits.

These checks verify that the reported scores are traceable to the original submission data and agree with the independent recomputation. Token and step counts come from the run telemetry; monetary cost uses the official leaderboard's \$15.209-per-task value.\cite{astabenchleaderboard}

\clearpage
\ifQiushiHasCitations
  \IfFileExists{refs_report.bib}{%
    \bibliographystyle{unsrt}
    \bibliography{refs_report}}{}
\fi

\end{document}